\documentclass[10pt,a4paper]{article}
\usepackage{a4wide}

\usepackage{hyperref}
\usepackage{url}

\usepackage[utf8]{inputenc} % allow utf-8 input
\usepackage[T1]{fontenc}    % use 8-bit T1 fonts
\usepackage{booktabs}       % professional-quality tables
\usepackage{amsfonts}       % blackboard math symbols
\usepackage{nicefrac}       % compact symbols for 1/2, etc.
\usepackage{microtype}      % microtypography

\usepackage{graphicx}
\usepackage{amsmath}
\usepackage{amssymb}
\usepackage{tabularx}
\usepackage{tikz,environ,pgfplots,booktabs}
\usepackage{amsthm}
\pgfplotsset{compat = 1.3}

\usepackage{soul}

\newcommand{\etal}{\mbox{\emph{et al.\ }}}
\newcommand{\abs}[1]{\left|#1\right|}
\newcommand{\norm}[1]{\left\|#1\right\|}

\def\Re{{\rm Re}}
\def\cC{\mathcal{C}}
\def\CC{\mathbb{C}}

\def\RR{\mathbb{R}}

\newtheorem{theorem}{Theorem}
\newtheorem{definition}[theorem]{Definition}

\newtheorem{remark}[theorem]{Remark}
\newtheorem{proposition}[theorem]{Proposition}

\newlength\fheight
\newlength\fwidth

\begin{document}

\title{CayleyNets: Graph Convolutional Neural Networks \\ with Complex Rational Spectral Filters}

\author{Ron~Levie$^*$,~
        Federico~Monti\thanks{The first two authors have contributed equally and are listed alphabetically.\newline
Ron Levie is with the Institute of Mathematics, Technische Universit{\"a}t Berlin, Berlin 10623, Germany (e-mail: levie@math.tu-berlin.de).\newline
Federico Monti is with the Institute of Computational Science, Universit\'a della Svizzera italiana, Lugano 6900, Switzerland (e-mail: federico.monti@usi.ch).\newline
Xavier Bresson is with the School of Computer Science and Engineering and the Data Science and AI Center at Nanyang Technological University (NTU), Singapore (email: xbresson@ntu.edu.sg). He is supported by NRF Fellowship NRFF2017-10.\newline
Michael M. Bronstein is with Imperial College London (UK), USI Lugano (Switzerland), and Intel Perceptual Computing (Israel).} ,~
				Xavier~Bresson~
        and~Michael~M.~Bronstein
 }

\date{ }

\maketitle

\begin{abstract}

The rise of graph-structured data such as social networks, regulatory networks, citation graphs, and functional brain networks, in combination with resounding success of deep learning in various applications, has brought the interest in generalizing deep learning models to non-Euclidean domains. In this paper, we introduce a new spectral domain convolutional architecture for deep learning on graphs. The core ingredient of our model is a new class of parametric rational complex functions (Cayley polynomials) allowing to efficiently compute spectral filters on graphs that specialize on frequency bands of interest. Our model generates rich spectral filters that are localized in space, scales linearly with the size of the input data for sparsely-connected graphs, and can handle different constructions of Laplacian operators. Extensive experimental results show the superior performance of our approach,  in comparison to other spectral domain convolutional architectures, on spectral image classification, community detection, vertex classification and matrix completion tasks.
\end{abstract}

\section{Introduction}

In many domains, one has to deal with large-scale data with underlying non-Euclidean structure. Prominent examples of such data are social networks, genetic regulatory networks, functional networks of the brain, and 3D shapes represented as discrete manifolds. 
The recent success of deep neural networks and, in particular, convolutional neural networks (CNNs) \cite{lecun1998gradient} have raised the interest in {\em geometric deep learning} techniques trying to extend these models to data residing on graphs and manifolds. 
 In this paper we focus on spectral graph CNNs.
Geometric deep learning approaches, { and specifically spectral graph CNNs,} have been successfully applied to computer graphics and vision  \cite{masci2015geodesic,WFT2015,add16,boscaini2016learning,monti2016geometric}, brain imaging \cite{ktena2017distance}, and drug design \cite{duv2015convolutional} problems, to mention a few. For a comprehensive presentation of methods and applications of deep learning on graphs and manifolds, we refer the reader to the review paper \cite{review_new}.

\subsection{Related work}
The earliest neural network formulation on graphs was proposed by \cite{gori2005new} and \cite{GNN}, combining random walks with recurrent neural networks (their paper has recently enjoyed renewed interest in \cite{GGSNN,comnets}). 
The first CNN-type architecture on graphs was proposed by \cite{bruna2013spectral}. One of the key challenges of extending CNNs to graphs is the lack of vector-space structure and shift-invariance making the classical notion of convolution elusive. Bruna \etal  formulated convolution-like operations in the spectral domain, using the graph Laplacian eigenbasis as an analogy of the Fourier transform  \cite{shuman2013emerging}.  
\cite{defferrard2016convolutional} proposed an efficient filtering scheme using recurrent Chebyshev polynomials applied on the Laplacian operator.  As opposed to \cite{bruna2013spectral}, Chebyshev filters are defined as functions $\mathbb{R}\rightarrow\mathbb{R}$ applied on the spectrum, as in \cite{shuman2013emerging}. This makes filters learned on one graph generalizable to other graphs. 
\cite{welling2016} simplified this architecture using filters operating on 1-hop neighborhoods of the graph. 
{\cite{atwood2016search} proposed a Diffusion CNN architecture based on powers of the degree-normalized transition matrix. }
\cite{monti2016geometric}  (and later, \cite{hechtlinger2017generalization}) proposed a spatial-domain generalization of CNNs to graphs using local patch operators represented as Gaussian mixture models, showing a significant advantage of such models in generalizing across different graphs. 
In \cite{monti2017geometric}, spectral graph CNNs were extended to multiple graphs and applied to matrix completion and recommender system problems.

\subsection{Main contribution}
In this paper, we construct graph CNNs employing an efficient spectral filtering scheme based on {the new class of} Cayley polynomials that enjoys similar advantages of the Chebyshev filters \cite{defferrard2016convolutional} such as localization and linear complexity {in the number of edges}. The main advantage of our filters over \cite{defferrard2016convolutional} is their ability to detect narrow frequency bands of importance during training, and to specialize on them while being well-localized on the graph. { We demonstrate experimentally that this affords our method greater flexibility, making it perform better than ChebNets on a broad range of graph learning problems. }

\subsection{Notation}
 We use $a, \mathbf{a}$, and $\mathbf{A}$ to denote scalars, vectors, and matrices, respectively. $\bar{z}$ denotes the conjugate of a complex number, $\mathrm{Re}\{z\}$ its real part, and $i=\sqrt{-1}$ denotes the imaginary unit.
$\mathrm{diag}(a_1, \hdots, a_n)$ denotes an $n\times n$ diagonal matrix with diagonal elements $a_1,\hdots, a_n$. 
$\mathrm{Diag}(\mathbf{A}) = \mathrm{diag}(a_{11}, \hdots, a_{nn})$ denotes an $n\times n$ diagonal matrix obtained by setting to zero the off-diagonal elements of $\mathbf{A}$. 
$\mathrm{Off}(\mathbf{A}) = \mathbf{A} - \mathrm{Diag}(\mathbf{A})$ denotes the matrix containing only the off-diagonal elements of $\mathbf{A}$. $\mathbf{I}$ is the identity matrix and 
$\mathbf{A}\circ \mathbf{B}$ denotes the Hadamard (element-wise) product of matrices $\mathbf{A}$ and $\mathbf{B}$. 
Proofs are given in the appendix.

\section{Spectral techniques for deep learning on graphs}
\label{sec:graph_cnn}
\subsection{Spectral graph theory}
Let $\mathcal{G} = (\{1,\hdots, n\}, \mathcal{E}, \mathbf{W})$ be an undirected weighted graph, represented by a symmetric {\em adjacency matrix} $\mathbf{W} = (w_{ij})$. We define $w_{ij} = 0$ if $(i,j) \notin \mathcal{E}$ and $w_{ij} > 0$ if $(i,j) \in \mathcal{E}$. 
We denote by $\mathcal{N}_{k,m}$ the {\em $k$-hop neighborhood} of vertex $m$, containing vertices that are at most $k$ edges away from $m$. 
The {\em unnormalized graph Laplacian} is an $n\times n$ symmetric positive-semidefinite matrix $\boldsymbol{\Delta}_u = \mathbf{D} - \mathbf{W}$, where $\mathbf{D} = \mathrm{diag}(\sum_{j\neq i} w_{ij} )$ is the {\em degree matrix}. 
The {\em normalized graph Laplacian} is defined as $\boldsymbol{\Delta}_\mathrm{n} = \mathbf{D}^{-1/2} \boldsymbol{\Delta}_u \mathbf{D}^{-1/2}  = \mathbf{I} - \mathbf{D}^{-1/2} \mathbf{W} \mathbf{D}^{-1/2}$. In the following, we use the generic notation $\boldsymbol{\Delta}$ to refer to some Laplacian.

Since both normalized and unnormalized Laplacian are symmetric and positive semi-definite matrices, they admit an eigendecomposition 
$
\boldsymbol{\Delta} = \boldsymbol{\Phi} \boldsymbol{\Lambda} \boldsymbol{\Phi}^\top
$, 
where $\boldsymbol{\Phi} = (\boldsymbol{\phi}_1, \hdots \boldsymbol{\phi}_n)$ are the orthonormal eigenvectors and $\boldsymbol{\Lambda} = \mathrm{diag}(\lambda_1, \hdots, \lambda_n)$ is the diagonal matrix of corresponding non-negative eigenvalues (spectrum) $0=\lambda_1 \leq \lambda_2 \leq \hdots \leq \lambda_n$. The eigenvectors play the role of Fourier atoms in classical harmonic analysis and the eigenvalues can be interpreted as (the square of) frequencies. 
Given a signal $\mathbf{f} = (f_1, \hdots, f_n)^\top$ on the vertices of graph $\mathcal{G}$, its {\em graph Fourier transform} is given by $\hat{\mathbf{f}} = \boldsymbol{\Phi}^\top\mathbf{f}$. 
Given two signals $\mathbf{f}, \mathbf{g}$ on the graph, their {\em spectral convolution} can be defined as the element-wise product of the  Fourier transforms, $\mathbf{f} \star \mathbf{g} = \boldsymbol{\Phi} \big( (\boldsymbol{\Phi}^\top\mathbf{g}) \circ (\boldsymbol{\Phi}^\top\mathbf{f}) \big) = \boldsymbol{\Phi}\, \mathrm{diag}(\hat{g}_1, \hdots, \hat{g}_n)\hat{\mathbf{f}},$
which corresponds to the property referred to as the {\em Convolution Theorem} in the Euclidean case. 

\subsection{Spectral CNNs}
\cite{bruna2013spectral} used the spectral definition of convolution %~(\ref{spectral_conv})
to generalize CNNs on graphs, with a spectral convolutional layer of the form 
\begin{equation} 
\label{spectral_construction_eq}
\mathbf{f}^{\mathrm{out}}_l =   \xi \left(  \sum_{l'=1}^{p} \boldsymbol{\Phi}_k \hat{\mathbf{G}}_{l,l'} \boldsymbol{\Phi}_k^\top \mathbf{f}^{\mathrm{in}}_{l'} \right).
\end{equation}
Here the $n\times p$ and $n\times q$ matrices $\mathbf{F}^{\mathrm{in}} = (\mathbf{f}^{\mathrm{in}}_1, \hdots, \mathbf{f}^{\mathrm{in}}_p)$  and $\mathbf{F}^{\mathrm{out}} = (\mathbf{f}^{\mathrm{out}}_1, \hdots, \mathbf{f}^{\mathrm{out}}_q)$ represent respectively the $p$- and $q$-dimensional input and output signals on the vertices of the graph, 
$\boldsymbol{\Phi}_k = (\boldsymbol{\phi}_1, \hdots, \boldsymbol{\phi}_k)$ is an $n\times k$ matrix of the first eigenvectors, 
$\hat{\mathbf{G}}_{l,l'} = \mathrm{diag}(\hat{g}_{l,l',1}, \hdots, \hat{g}_{l,l',k})$ is a $k\times k$ diagonal matrix of spectral multipliers representing a learnable filter in the frequency domain, and $\xi$ is a nonlinearity (e.g., ReLU) applied on the vertex-wise function values. 
Pooling is performed by means of graph coarsening, which, given a graph with $n$ vertices, produces a graph with $n'<n$ vertices and transfers signals from the vertices of the fine graph to those of the coarse one. {Assuming $k = \mathcal{O}(n)$ Laplacian eigenvectors are used, a spectral convolutional layer requires $\mathcal{O}(p q k) = \mathcal{O}(n)$ parameters to train.} 
{
In addition, an informal approach for keeping the filters localized in the spectral domain was proposed. The idea is to learn just a few spectral coefficients of the filter, and obtain the rest using interpolation. This also keeps the number of filter parameters $\mathcal{O}(1)$. The spatial locality property simulates local receptive fields \cite{pro:CoatesNg11LRF}, and is important for the interpretability of convolutions as filters. Moreover, for spatial implementation of (\ref{spectral_construction_eq}), such as ChebNet and CayleyNet, small receptive fields typically indicate sparse implementations. }

This framework has several major drawbacks. 
First, the computation of the forward and inverse graph Fourier transforms incur expensive $\mathcal{O}(n^2)$ multiplication by the matrices $\boldsymbol{\Phi}, \boldsymbol{\Phi}^\top$, as there is no FFT-like algorithms on general graphs.

Second, the spectral filter coefficients are {\em basis dependent}, and consequently, a spectral CNN model learned on one graph cannot be transferred %applied
 to another graph.
{ This is as opposed to \cite{shuman2013emerging}, where the frequency responses of the filters coefficients are represented as $\hat{g}_i = g(\lambda_i)$, where $g(\lambda)$ is a smooth transfer function of frequency $\lambda$. Applying such filter to signal $\mathbf{f}$ can be expressed as $\mathbf{G}\mathbf{f} = g(\boldsymbol{\Delta})\mathbf{f}  = \boldsymbol{\Phi} g(\boldsymbol{\Lambda}) \boldsymbol{\Phi}^\top \mathbf{f} = \boldsymbol{\Phi} \,\mathrm{diag}(g(\lambda_1), \hdots, g(\lambda_n)) \boldsymbol{\Phi}^\top \mathbf{f}$, where applying a function to a matrix is understood in the operator functional calculus sense (applying the function to the matrix eigenvalues).}

{ It is noteworthy to mention alternative functional calculus driven approaches to define convolution. In \cite{art:Sandryhaila1} filters are defined as functions of the adjacency matrix $g(\mathbf{W})$, and in \cite{art:Sandryhaila2} the problem of ordering the eigenvalues of $\mathbf{W}$ according to a natural notion of frequency was addressed.}

\subsection{ChebNet}
\cite{defferrard2016convolutional}  used polynomial filters represented in the Chebyshev basis 
\begin{equation} \label{eq:filt_cheby}
	g_{\boldsymbol{\alpha}}(\tilde{\lambda}) = \sum_{j=0}^{r} \alpha_j T_j(\tilde{\lambda}) \vspace{-0.5mm}
\end{equation}
applied to rescaled frequency $\tilde{\lambda} \in [-1,1]$; here, $\boldsymbol{\alpha}$ is the $(r+1)$-dimensional vector of polynomial coefficients parametrizing the filter and optimized for during the training, and $T_j(\lambda) = 2\lambda T_{j-1}(\lambda) - T_{j-2}(\lambda)$ denotes the Chebyshev polynomial of degree $j$ defined in a recursive manner with $T_1(\lambda) =\lambda$ and $T_0(\lambda) =1$. Chebyshev polynomials form an orthogonal basis for the space of polynomials of order $r$ on $[-1,1]$.
Applying the filter is performed by $g_{\boldsymbol{\alpha}}(\tilde{\boldsymbol{\Delta}}) \mathbf{f}$, where $\tilde{\boldsymbol{\Delta}} = 2 \lambda_{n}^{-1}\boldsymbol{\Delta}  - \mathbf{I}$ is the rescaled Laplacian such that its eigenvalues $\tilde{\boldsymbol{\Lambda}} = 2 \lambda_{n}^{-1} \boldsymbol{\Lambda}  - \mathbf{I}$ are in the interval $[-1,1]$.

Such an approach has several important advantages. 
First, since 
$g_{\boldsymbol{\alpha}}(\tilde{\boldsymbol{\Delta}}) = \sum_{j=0}^r \alpha_j T_j(\tilde{\boldsymbol{\Delta}})$ contains only matrix powers, additions, and multiplications by scalar, it can be computed avoiding the explicit expensive $\mathcal{O}(n^3)$ computation of the Laplacian eigenvectors.  
Furthermore, due to the recursive definition of the Chebyshev polynomials, the computation of the filter $g_{\boldsymbol{\alpha}}(\boldsymbol{\Delta}) \mathbf{f}$ entails applying the Laplacian $r$ times, resulting in $\mathcal{O}(rn)$ operations assuming that the Laplacian is a sparse matrix with $\mathcal{O}(1)$ non-zero elements in each row (a valid hypothesis for most real-world graphs that are sparsely connected). {From another point of view, in each multiplication by the Laplacian, neighbors in the graph exchange data, and there are overall $r$ such neighbor exchanges.}
Second, the number of parameters is $\mathcal{O}(1)$ as $r$ is independent of the graph size $n$. 
Third, since the Laplacian is a local operator affecting only 1-hop neighbors of a vertex and a polynomial of degree $r$ of the Laplacian affects only $r$-hops, the resulting filters have guaranteed spatial localization.

A key disadvantage of Chebyshev filters is the fact that using polynomials makes it hard to produce narrow-band filters, as such filters require very high order $r$, and produce unwanted non-local filters. This deficiency is especially pronounced when the Laplacian has clusters of eigenvalues concentrated around a few frequencies with large spectral gap (Figure~\ref{fig:perf-multiple-orders}, second to the last). { Indeed, in ChebNets the Laplacian eigenvalues are contracted to the band $[-1,1]$, and the clusters of eigenvalues become very concentrated. Now, the frequency response of the Chebyshev filter is a polynomial in $[-1,1]$, which is unable to separate the individual eigenvalues in the clusters due to an uncertainty principle.} Such a behavior is characteristic of graphs with community structures, which is very common in many real-world graphs, for instance, social networks. 

{ Let us explain the above phenomenon more accurately. Recall that Chebyshev polynomials are given by
\[T_n(\cos(\theta))=\cos(n\theta),\]
and form an orthonormal basis of the weighted Lebesgue space $L_2\big([-1,1],\frac{1}{\sqrt{1-x^2}}dx\big)$.
When the variable is changed via $x=\cos(\theta)$, the Chebyshev basis maps to the cosine basis in $[0,\pi]$, and the space $L_2\big([-1,1],\frac{1}{\sqrt{1-x^2}}dx\big)$ maps to the space $L_2(0,\pi)$ with the standard Lebesgue measure. Now, suppose that we want to represent a band-pass filter on the narrow band $[\cos(b),\cos(a)]\subset [-1,1]$, with small $\cos(b-a)$. Under the change of variable, this band maps to $[a,b]$ with small $b-a=\epsilon$. In this case, since the characteristic function of $[a,b]$ is the shrinking dilation of the characteristic function of $[-\frac{1}{2},\frac{1}{2}]$ (up to translation), it's Fourier coefficients are samples from the stretching dilation of the Fourier transform of the characteristic function of $[-\frac{1}{2},\frac{1}{2}]$. As a result, the number of coefficients required to approximate a band pass filter up to some fixed tolerance is inverse proportional to the size of the band. More generally, the number of Chebyshev coefficients required for approximating a filter having features in a given scale, is inverse proportional to the scale.
When the Laplacian has a cluster of eigenvalues concentrated around one frequency, a filter that separates these eigenvalues must have features in scale proportional to the radius of the eigenvalue cluster. Therefore, the number of Chebyshev coefficients must be inverse proportional to the cluster size.}

To overcome this major drawback, we need a new class of filters, {that both entail $\mathcal{O}(r)$ neighbor exchanges, and are able to specialize in narrow bands in frequency.}

\section{Cayley filters}
A key construction of this paper is a family of complex filters that enjoy the advantages of Chebyshev filters while avoiding some of their drawbacks. 
 { We define} a {\em Cayley polynomial} of order  $r$ to be a real-valued function with complex coefficients, \vspace{-1mm}%$\mathbf{G}$ applied to a signal $\mathbf{f}$ 
\begin{equation}
\label{eq:cayleyf}
%\mathbf{G}\mathbf{f} 
g_{\mathbf{c},h}(\lambda)  = c_0  + 2\Re \Big\{ \sum_{j=1}^r c_j(h\lambda-i)^j(h\lambda+i)^{-j} \Big\}
\end{equation}
where $\mathbf{c} = (c_0, \hdots, c_r)$ is a vector of one real coefficient and $r$ complex coefficients and $h>0$ is the {\em spectral zoom} parameter, that will be discussed later. % and $\Re\{\mathbf{z}\}$ denotes the real part of the complex vector $\mathbf{z}$. 
A {\em Cayley filter} $\mathbf{G}$ is a spectral filter %based on a Cayley polynomial, 
defined on real signals $\mathbf{f}$ by  \vspace{-1mm}
\begin{equation}
\label{eq:cayleyf1}
\mathbf{G}\mathbf{f} = 
g_{\mathbf{c},h}(\boldsymbol{\Delta}) \mathbf{f}  = c_0 \mathbf{f} + 2\Re \{ \sum_{j=1}^r c_j(h\boldsymbol{\Delta}-i\mathbf{I})^j(h\boldsymbol{\Delta}+i\mathbf{I})^{-j} \mathbf{f} \}, 
\end{equation}
%
%p
where the parameters $\mathbf{c}$ and $h$ are {optimized during training}. 
Similarly to the Chebyshev filters, Cayley filters involve basic matrix operations such as powers, additions, multiplications by scalars, and also inversions. This implies that application of the filter $\mathbf{G}\mathbf{f}$ can be performed without explicit expensive eigendecomposition of the Laplacian operator.  Cayley filters are special cases of filters based on general rational functions of the Laplacian, namely ARMA filters \cite{art:ARMA}\cite{art:filter_ARMA}. For a general rational functions of the Laplacian, calculating the denominator requires a matrix inversion. When the filter is based on arbitrary coefficients, there is no guarantee that the matrix inversions are calculated stably. Guaranteeing stable inversions for arbitrary filter coefficients is important, since the coefficients follow an unknown path during training. { For general ARMA filters, the filter can acquire poles arbitrarily close to the spectrum of $\boldsymbol{\Delta}$ during training, so there is no uniform analysis of convergence of the approximate inversions in the filter computation.
The motivation to use the subclass of Cayley filters over general rational functions is to guarantee that inversion is uniformly stable. Namely, the number of iterations required for a given approximation error is independent of the filter coefficients, as long as the coefficients are bounded.   
Moreover, the lower number of parameters in Cayley polynomials in comparison to general rational functions may be beneficial for avoiding overfitting.}

In the following, we show that Cayley filters are analytically well behaved; in particular, any smooth spectral filter can be represented as a Cayley polynomial, and low-order filters are localized in the spatial domain. We also discuss numerical implementation and compare Cayley and Chebyshev filters. { We show that Cayley filters defined on sparse Laplacians with $\mathcal{O}(1)$ non-zero elements takes $\mathcal{O}(n)$ operations, similarly to Chebyshev filters.}

\subsection{Analytic properties}
Cayley filters are best understood through the  Cayley transform, from which their name derives. 
Denote by $e^{i\RR}=\{e^{i\theta} : \theta\in\RR\}$ the unit complex circle.
The {\em Cayley transform} $\cC(x) = \frac{x-i}{x+i}$ is a 
smooth bijection between $\RR$ and $e^{i\RR}\setminus \{1\}$. 
The complex matrix 
$\cC(h\boldsymbol{\Delta})=(h\boldsymbol{\Delta}-i\mathbf{I})(h\boldsymbol{\Delta}+i\mathbf{I})^{-1}$ obtained by applying the Cayley transform to the scaled Laplacian $h\boldsymbol{\Delta}$ has its spectrum in $e^{i\RR}$ and is thus unitary. 
Since $z^{-1}=\overline{z}$ for $z\in e^{i\RR}$, we can write $\overline{c_j \cC^{j}(h \boldsymbol{\Delta})}=\overline{c_j} \cC^{-j}(h \boldsymbol{\Delta})$.
Therefore, using $2\Re\{z\} = z + \overline{z}$, any Cayley filter~(\ref{eq:cayleyf1}) 
can be written as a conjugate-even Laurent polynomial w.r.t. $\cC(h\boldsymbol{\Delta})$, \vspace{-1mm}
\begin{equation}
\mathbf{G}=c_0 \mathbf{I}+\sum_{j=1}^r c_j \cC^j(h \boldsymbol{\Delta})+ \overline{c_j} \cC^{-j}(h \boldsymbol{\Delta}). 
\label{eq:Cayley_filt_symm}
\end{equation}
Since the spectrum of $\cC(h\boldsymbol{\Delta})$ is in $e^{i\RR}$, the operator $\cC^j(h\boldsymbol{\Delta})$ can be thought of as a multiplication by a pure harmonic in the frequency domain $e^{i\RR}$ for any integer power $j$, \vspace{-1mm} 
\[\cC^j(h\boldsymbol{\Delta}) = \boldsymbol{\Phi}\mathrm{diag}\big(\big[\cC(h\lambda_1)\big]^j, \hdots, \big[\cC(h\lambda_n)\big]^j\big)\boldsymbol{\Phi}^\top. \]
A Cayley filter can be thus  seen as a multiplication by a finite Fourier expansions in the frequency domain $e^{i\RR}$. 
Since (\ref{eq:Cayley_filt_symm}) is conjugate-even, it is a (real-valued) trigonometric polynomial.

Note that any spectral filter can be formulated as a Cayley filter. Indeed, spectral filters $g(\boldsymbol{\Delta})$ are specified by the finite sequence of values $g(\lambda_1), \hdots, g(\lambda_n)$, which can be interpolated by a trigonometric polynomial. Moreover, since trigonometric polynomials are smooth, we expect low order Cayley filters to be well localized in some sense on the graph, as discussed later. 

Finally, in definition (\ref{eq:cayleyf1}) we use complex coefficients. If $c_j\in \RR$ then (\ref{eq:Cayley_filt_symm}) is an even cosine polynomial, and if $c_j\in i\RR$ then (\ref{eq:Cayley_filt_symm}) is an odd sine polynomial. Since the spectrum of $h\boldsymbol{\Delta}$ is in $\RR_+$, it is mapped to the lower half-circle by $\cC$, on which both cosine and sine polynomials are complete and can represent any spectral filter. However, it is beneficial to use general complex coefficients, since complex Fourier expansions are overcomplete in the lower half-circle, thus describing a larger variety of spectral filters of the same order without increasing the computational complexity of the filter. 

\subsection{Spectral zoom}
\begin{figure}[!ht]
\vspace{0mm}
\centering 
\includegraphics[width=1\linewidth]{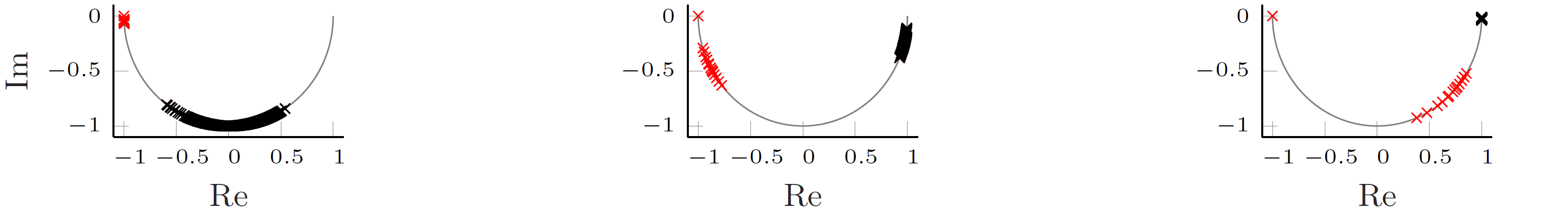}
\vspace{-4mm}
\caption{Eigenvalues of the unnormalized Laplacian $h\boldsymbol{\Delta}_\mathrm{u}$ of the 15-communities graph mapped on the complex unit half-circle by means of Cayley transform with spectral zoom values (left-to-right) $h=0.1$, $1$, and $10$. The first 15 frequencies carrying most of the information about the communities are marked in red. Larger values of $h$ zoom (right) on the low frequency band.}\vspace{-1mm}
\label{fig:circle}
\end{figure}

To understand the essential role of the parameter $h$ in the Cayley filter, consider $\cC(h\boldsymbol{\Delta})$. Multiplying $\boldsymbol{\Delta}$ by $h$ dilates its spectrum, and applying $\cC$ on the result maps the non-negative spectrum to the complex half-circle.  
The greater $h$ is, the more the spectrum of $h\boldsymbol{\Delta}$ is spread apart in $\RR_+$, resulting in better spacing of the smaller eigenvalues of $\cC(h\boldsymbol{\Delta})$. On the other hand,
the smaller $h$ is, the further away the high frequencies of $h\boldsymbol{\Delta}$ are from $\infty$, the better spread apart are the high frequencies of $\cC(h\boldsymbol{\Delta})$ in $e^{i\RR}$ (see Figure~\ref{fig:circle}).  
Tuning the parameter $h$ allows thus to `zoom' in to different parts of the spectrum, resulting in filters specialized in different frequency bands.

%SLOW
%\begin{figure}[!ht]
%\vspace{0mm}
%\centering 
%\center
%\setlength\fheight{1.3cm}
%\setlength\fwidth{0.16\linewidth}
%\input{figures/plot_mapped_eigenvalues/quasi-disconnected_graph/plot_h=0.10.tikz}\hfill
%\input{figures/plot_mapped_eigenvalues/quasi-disconnected_graph/plot_h=1.002.tikz}\hfill
%\input{figures/plot_mapped_eigenvalues/quasi-disconnected_graph/plot_h=10.002.tikz}\vspace{-2mm}
%\caption{Eigenvalues of the unnormalized Laplacian $h\boldsymbol{\Delta}_\mathrm{u}$ of the 15-communities graph mapped on the complex unit half-circle by means of Cayley transform with spectral zoom values (left-to-right) $h=0.1$, $1$, and $10$. The first 15 frequencies carrying most of the information about the communities are marked in red. Larger values of $h$ zoom (right) on the low frequency band.}\vspace{-4mm}
%\label{fig:circle}
%\end{figure}

\subsection{Numerical properties}
\label{Numerical properties}
The numerical core of the Cayley filter is the computation of 
$\cC^j(h\boldsymbol{\Delta}) \mathbf{f}$ for $j=1,\ldots,r$, performed in a sequential manner.
Let $\mathbf{y}_0,\ldots,\mathbf{y}_r$ denote the solutions of the following linear recursive system, 
\begin{equation}
\mathbf{y}_0=\mathbf{f}, \quad \quad (h\boldsymbol{\Delta}+i\mathbf{I})\mathbf{y}_j=(h\boldsymbol{\Delta}-i\mathbf{I}) \mathbf{y}_{j-1} \ ,\ j=1,\ldots,r.
\label{eq:y_array}
\end{equation}
Note that sequentially approximating $\mathbf{y}_j$ in (\ref{eq:y_array}) using the approximation of $\mathbf{y}_{j-1}$ in the {right hand side} is stable, since $\cC(h\boldsymbol{\Delta})$ is unitary and thus has condition number $1$.

Equations (\ref{eq:y_array}) can be solved with matrix inversion exactly, but it costs $\mathcal{O}(n^3)$. An alternative is to use the Jacobi method,\footnote{We remind that the Jacobi method for solving $\mathbf{A}\mathbf{x} = \mathbf{b}$  consists in decomposing $\mathbf{A} = \mathrm{Diag}(\mathbf{A}) + \mathrm{Off}(\mathbf{A})$ and obtaining the solution iteratively as 
$\mathbf{x}^{(k+1)} = -( \mathrm{Diag}(\mathbf{A}))^{-1} \mathrm{Off}(\mathbf{A})\mathbf{x}^{(k)} +   ( \mathrm{Diag}(\mathbf{A}))^{-1}\mathbf{b}$.
} 
which provides approximate solutions  $\tilde{\mathbf{y}}_j\approx \mathbf{y}_j$. 
Let $\mathbf{J} = -(\mathrm{Diag}(h\boldsymbol{\Delta}+ i\mathbf{I}))^{-1}\mathrm{Off}(h\boldsymbol{\Delta}+ i\mathbf{I})$ be the Jacobi iteration matrix associated with equation~(\ref{eq:y_array}).  
For the unnormalized Laplacian,  
$\mathbf{J}=(h\mathbf{D}+i\mathbf{I})^{-1}h\mathbf{W}$.
Jacobi iterations for approximating (\ref{eq:y_array}) for a given $j$ have the form
\begin{equation}
\begin{split}
\tilde{\mathbf{y}}_j^{(k+1)} =& \mathbf{J} \tilde{\mathbf{y}}_j^{(k)} + \mathbf{b}_j\\
 \mathbf{b}_j= &(\mathrm{Diag}(h\boldsymbol{\Delta}+ i\mathbf{I}))^{-1}(h\boldsymbol{\Delta}-i\mathbf{I}) \tilde{\mathbf{y}}_{j-1},
\end{split}
\label{eq:Jaco}
\end{equation}
initialized with $\tilde{\mathbf{y}}_j^{(0)}=\mathbf{b}_j$ and terminated after $K$ iterations, yielding
$\tilde{\mathbf{y}}_j = \tilde{\mathbf{y}}_j^{(K)}$. 
We denote $\tilde{\mathbf{y}}_0={\mathbf{y}}_0$.
The application of the approximate Cayley filter is given by 
$\widetilde{\mathbf{G}\mathbf{f}} = { c_0 \tilde{\mathbf{y}}_0 +2 {\rm Re}}\sum_{j=1}^r c_j \tilde{\mathbf{y}}_j \approx \mathbf{G}\mathbf{f}$, and takes $\mathcal{O}(rKn)$ operations under the previous assumption of a sparse Laplacian. 
The method can be improved by normalizing $\norm{\tilde{\mathbf{y}}_{j}}_2=\norm{\mathbf{f}}_2$.

Next, we give an error bound for the approximate filter.   
For the unnormalized Laplacian, let $d=\max_{j}\{d_{j,j}\}$ and $\kappa =\norm{\mathbf{J}}_{\infty}= \frac{hd}{\sqrt{h^2d^2+1}} <1$. 
For the normalized Laplacian, we assume that $(h\boldsymbol{\Delta}_n+i\mathbf{I})$ is dominant diagonal, which gives $\kappa=\norm{\mathbf{J}}_{\infty}<1$.

\begin{proposition}
\label{error}
Under the above assumptions, \newline
$\frac{\| \mathbf{G}\mathbf{f}-\widetilde{\mathbf{G}\mathbf{f}}\|_2}{\norm{\mathbf{f}}_2} \leq  2M \kappa^K$,
where
$M=\sqrt{n}\sum_{j=1}^rj\abs{c_j}$ in the general case, and $M=\sum_{j=1}^rj\abs{c_j}$ if the graph is regular.
\end{proposition} 
Proposition \ref{error} is pessimistic in the general case, while requires strong assumptions in the regular case. We find that in most real life situations the behavior is closer to the regular case.
It also follows from Proposition \ref{error} that 
smaller values of the spectral zoom $h$ result in faster convergence, giving this parameter an additional numerical role of accelerating convergence. 

Last, note that a Cayley filter with Jacobi approximation, is based on powers of the Jacobi matrix ${\mathbf{J}}$, in addition to $(h\boldsymbol{\Delta}-i\mathbf{I})$. The Jacobi matrix can be viewed as a general representation matrix of the graph, replacing the standard Laplacian with a matrix that respects the connectivity of the graph (general representation matrices were considered in \cite{art:ARMA}). This is also true for $(h\boldsymbol{\Delta}-i\mathbf{I})$.  In this point of view, learning $h$ is interpreted as learning a general representation matrix. Learning $h$ can be also viewed as learning a normalization of the weights of the graph. The problem of learning the topology of the graph was studied e.g. in \cite{henaff2015deep}.

\subsection{Complexity}
In practice, an accurate inversion of $(h\boldsymbol{\Delta}+i\mathbf{I})$ is not required, since the approximate inverse is combined with learned coefficients, which “compensate”, as necessary, for the inversion inaccuracy. 
{ Such behavior is well-documented in the literature in other contexts of model compression and accelerated convergence of iterative algorithms. For example, in \cite{Gregor:2010}, sparse signal coding are learned by unrolling iterative shrinkage algorithms (FISTA) into a neural network, where each layer emulates an iteration of the original algorithm but has extra learnable parameters. It is shown that FISTA networks with just a few layers outperform hundreds or thousands of iterations of the original algorithm thanks to the learnable parameters. The above phenomenon is also a common observation for solving sparse linear equations for compressed sensing tasks (see e.g \cite{YinError}).}

In a CayleyNet for a fixed graph, we fix the number of Jacobi iterations. Since the convergence rate depends on $\kappa$, that depends on the graph, different graphs may need different numbers of iterations. The convergence rate also depends on $h$.
Since there is a trade-off between the spectral zoom amount $h$, and the accuracy of the approximate inversion, and since $h$ is a learnable parameter, the training finds the right balance between the spectral zoom amount and the inversion accuracy.

{ To formulate computational complexity results, we consider the case where the number of vertices $n$ is ``big''. To formalize this, we consider a sequence of graphs indexed by $n$, and study the asymptotics as $n\rightarrow\infty$. When the graph is sampled from a continuous entity, like a manifold, this asymptotic analysis has a precise meaning. Otherwise, the asymptotic analysis is just a formal way of saying ``big $n$''.}
%We study the computational complexity of our method,  as the number of vertices $n$ of the graph tends to infinity. 
For every constant of a graph, e.g $d,\kappa$, we add the subscript $n$, indicating the number of vertices of the graph. 
We assume that there is a global constant $C$, such that the number of edges is bounded by $Cn$.
For the unnormalized Laplacian, we assume that $d_n$ and $h_n$ are bounded, which gives $\kappa_n<a<1$ for some $a$ independent of $n$. For the normalized Laplacian, we assume that $\kappa_n<a<1$. { These assumptions pose regularity on the sequence of graphs.} By {Proposition \ref{error}}, fixing the number of Jacobi iterations $K$ and the order of the filter $r$, independently of $n$, keeps the Jacobi error controlled. As a result, the number of parameters of the Cayley filters can be kept $\mathcal{O}(1)$, and for a Laplacian modeled as a sparse matrix, applying a Cayley filter on a signal takes $\mathcal{O}(n)$ operations. {Indeed, the Jacobi matrix $\mathbf{J}$ has the same connectivity as the graph, including edges connecting each vertex to itself. In each Jacobi iteration, $\mathbf{J}$ is applied $K$ times, which means that vertices exchange information with their neighbors $K$ times, and one time in the initialization due to one multiplication by $(h\boldsymbol{\Delta}-iI)$. The Jacobi approximation of ${\cal C}(h{\boldsymbol{\Delta}})$ is computed $r$ times, for the $r$ coefficients of the filter, and thus overall there are $(K+1)r$ neighbor exchanges in the method. Thus, for a sparsely connected graph with $\mathcal{O}(n)$ edges, applying a Cayley filter on a signal takes $\mathcal{O}((K+1)rn)$ operations.}

\begin{figure}[!ht]
\vspace{-1mm}
\centering 
\includegraphics[width=1\linewidth]{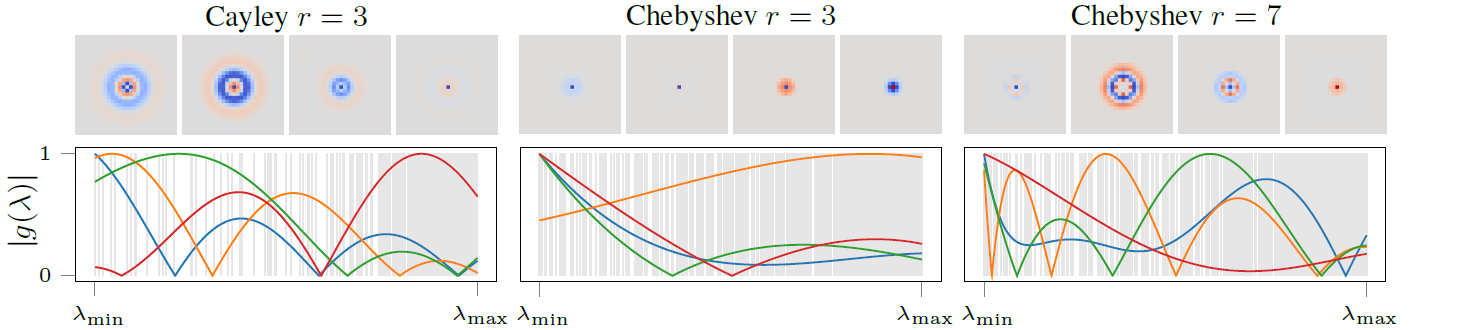}
\vspace{-3mm}
\caption{Filters (spatial domain, top and spectral domain, bottom) learned by CayleyNet (left) and ChebNet (center, right) on the MNIST dataset. Cayley filters are able to realize larger supports for the same order $r$.}\vspace{-1mm}
\label{fig:MNIST-filters}
\end{figure}

\subsection{Localization}
Unlike Chebyshev filters that have the small $r$-hop support, 
Cayley filters are rational functions supported on the whole graph. 
However, it is still true that Cayley filters are well localized on the graph.  
Let $\mathbf{G}$ be a Cayley filter and
 $\boldsymbol{\delta}_m$ denote a delta-function on the graph, defined as one at vertex $m$ and zero elsewhere. We show that $\mathbf{G} \boldsymbol{\delta}_m$ decays fast, in the following sense:

\begin{definition}[Exponential decay on graphs]
\label{exp_decay}
Let $\mathbf{f}$ be a signal on the vertices of graph $\mathcal{G}$, $1\leq p\leq \infty$, and $0<\epsilon<1$.
	Denote by $S \subseteq \{1, \hdots, n\}$ a subset of the vertices and by $S^c$ its complement. 
	We say that the {\em $L_p$-mass of $\mathbf{f}$ is supported in $S$ up to $\epsilon$} if 
	$\norm{\mathbf{f} |_{S^c}}_p \leq \epsilon\norm{\mathbf{f}}_p$, where $\mathbf{f} |_{S^c} = (f_{l})_{l \in S^c}$ is the restriction of $\mathbf{f}$ to $S^c$.
	We say that $\mathbf{f}$ has (graph) {\em exponential decay about vertex $m$}, if there exists some $\gamma\in (0,1)$ and $c>0$ such that for any $k$, the $L_p$-mass of $\mathbf{f}$ is supported in ${\cal N}_{k,m}$ up to $c\gamma^k$. 
	Here,  ${\cal N}_{k,m}$ is the $k$-hop neighborhood of $m$.
\end{definition}

\begin{remark}
Note that Definition~\ref{exp_decay} is analogous to classical exponential decay on Euclidean space: $\abs{f(x)}\leq R \gamma^{-x}$ iff for every ball $B_{\rho}$ of radius $\rho$ about $0$, $\| f|_{B_{\rho}^c} \|_{\infty}\leq c\gamma^{-\rho}\norm{f}_{\infty}$ with $c=\frac{R}{\norm{f}_{\infty}}$.
\end{remark}

\begin{theorem}
\label{decay_theorem}
Let $\mathbf{G}$ be a Cayley filter of order $r$. 
Then, 
$\mathbf{G}\boldsymbol{\delta}_m$ has exponential decay about $m$ in $L_{2}$, with constants 
$c=4M\frac{1}{\|\mathbf{G}\boldsymbol{\delta}_m\|_2}$ and $\gamma=\kappa^{1/r}$ (where $M$ and $\kappa$ are from Proposition \ref{error}). 
\end{theorem}

\subsection{Cayley vs Chebyshev}
Below, we compare the two classes of filters:\\
\textcolor{black}{{\it Chebyshev as a special case of Cayley.} For a regular graph with $\mathbf{D}=d\mathbf{I}$, using Jacobi inversion based on {zero iteration}, we get that any Cayley filter of order $r$ is a polynomial of $\boldsymbol{\Delta}$ in the monomial base $\big(\frac{h\boldsymbol{\Delta}-i}{hd+i}\big)^j$. In this situation, a Chebyshev filter, which is a real valued polynomial of $\boldsymbol{\Delta}$, is a special case of a Cayley filter.}
\\
{\it Spectral zoom and stability.} Generally, both Chebyshev polynomials and trigonometric polynomials give stable approximations, optimal for smooth functions. However, this crude statement is over-simplified. 
One of the drawbacks in Chebyshev filters is the fact that the spectrum of $\boldsymbol{\Delta}$ is always mapped to $[-1,1]$ in a linear manner, making it hard to specialize in small frequency bands. In Cayley filters, this problem is mitigated with the help of the spectral zoom parameter $h$. As an example, consider the community detection problem discussed in the next section. A graph with strong communities has a cluster of small eigenvalues near zero. 
Ideal filters $g(\boldsymbol{\Delta})$ for extracting the community information should be able to focus on this band of frequencies. 
Approximating such filters with Cayley polynomials, we zoom in to the band of interest by choosing the right $h$, and then project $g$ onto the space of trigonometric polynomials of order $r$, getting a good and stable approximation (Figure~\ref{fig:perf-multiple-orders}, bottom). 
However, if we project $g$ onto the space of Chebyshev polynomials of order $r$, the interesting part of $g$  concentrated on a small band is smoothed out and lost (Figure~\ref{fig:perf-multiple-orders}, second to the last). Thus, projections are not the right way to approximate such filters, and the stability of orthogonal polynomials cannot be invoked. { On the other hand, if we want to approximate $g$ on the small band using polynomials, ignoring the behavior away from this band, the approximation will be unstable away from this band; small perturbations in $g$ will result in big perturbations in the Chebyshev filter away from the band. This is due to the fact that any polynomial diverges at infinity, and for an asymptotically small band and polynomials of fixed order, ``away from the band behaves like infinity.''} For this reason, we say that Cayley filters are more stable than Chebyshev filters.\vspace{0.5mm}\\
{\it Regularity.} We found that in practice, low-order Cayley filters are able to model both very concentrated impulse-like filters, and wider Gabor-like filters. Cayley filters are able to achieve a wider range of filter supports with less coefficients than Chebyshev filters (Figure \ref{fig:MNIST-filters}), making the Cayley class more regular than Chebyshev. \vspace{0.5mm}\\
{\it Complexity.} Under the assumption of sparse Laplacians, both Cayley and Chebyshev filters incur linear complexity $\mathcal{O}(n)$. Besides, the new filters are equally simple to implement as Chebyshev filters; as seen in Eq.\ref{eq:Jaco}, they boil down to simple sparse matrix-vector multiplications providing a GPU friendly implementation.

\section{Results}

\subsection{Experimental settings} 
We test the proposed CayleyNets reproducing the experiments of  
\cite{defferrard2016convolutional,welling2016,monti2016geometric} and using ChebNet \cite{defferrard2016convolutional} as our main baseline method. { Pooling and graph coarsening was performed identically to \cite{defferrard2016convolutional}. The hyperparameters are identical to the original experiments, and not optimized.}
 All the methods were implemented in TensorFlow \cite{art:Tensorflow16}. The experiments were executed on a machine with a 3.5GHz Intel Core i7 CPU, 64GB of RAM, and NVIDIA Titan X GPU with 12GB of RAM.  SGD+Momentum and Adam \cite{KingmaB14} optimization methods were used to train the models in MNIST and the rest of the experiments, respectively. 
 Training and testing were always done on disjoint sets.

\subsection{Community detection}
\begin{figure}[!ht]
\vspace{-1mm}
\centering 
\includegraphics[width=1\linewidth]{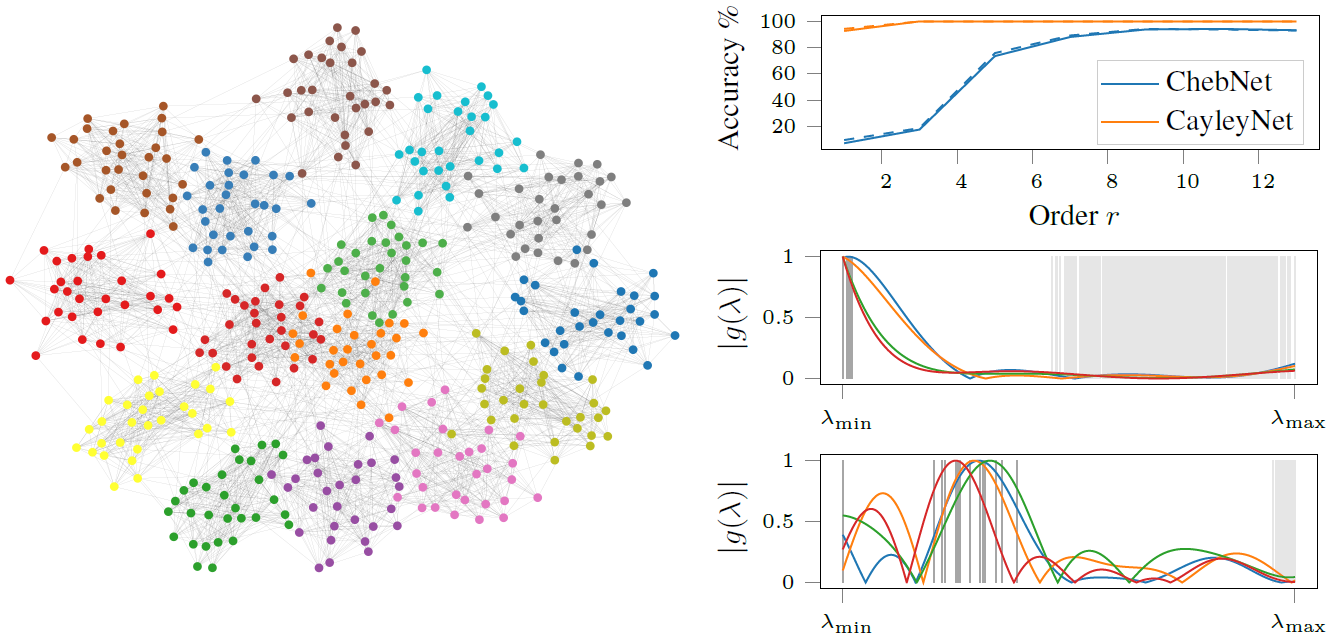}
\vspace{-3mm}
\caption{Left: synthetic 15-communities graph. Right: community detection accuracy of ChebNet and CayleyNet (top);  normalized responses of four different filters learned by ChebNet (middle) and CayleyNet (bottom). Grey vertical lines represent the frequencies of the normalized Laplacian ($\tilde{\lambda} = 2\lambda_n^{-1} \lambda - 1$ for ChebNet and $C(\lambda) = (h\lambda - i)/(h\lambda + i)$ unrolled to a real line for CayleyNet). Note how thanks to spectral zoom property Cayley filters can focus on the band of low frequencies (dark grey lines) containing most of the information about communities.}
\label{fig:perf-multiple-orders}
\end{figure}

We start with an experiment on a synthetic graph consisting of 15 communities with strong connectivity within each community and sparse connectivity across communities (Figure \ref{fig:perf-multiple-orders}, top). Though rather simple, such a dataset allows to study the behavior of different algorithms in controlled settings. 
On this graph, we generate noisy step signals, defined as $f_i = 1 + \sigma_i$ if $i$ belongs to the community, and $f_i = \sigma_i$ otherwise, where $\sigma_i \sim \mathcal{N}(0,0.3)$ is Gaussian i.i.d. noise. 
The goal is to classify each such signal according to the community it belongs to. The neural network architecture used for this task consisted of a spectral convolutional layer (based on Chebyshev or Cayley filters) with 32 output features, a mean pooling layer, and a softmax classifier for producing the final classification into one of the 15 classes. { No regularization has been exploited in this setting.} The classification accuracy is shown in Figure \ref{fig:perf-multiple-orders} (second to the top) along with examples of learned filters (bottom two). We observe that CayleyNet significantly outperforms ChebNet for smaller filter orders, with an improvement as large as 80\%. 
Studying the filter responses, we note that due to the capability to learn the spectral zoom parameter, CayleyNet allows to generate band-pass filters in the low-frequency band that discriminate well the communities (Figure \ref{fig:perf-multiple-orders} bottom). 

%SLOW
%\begin{figure}[!ht]
%\vspace{-1mm}
%\centering 
%\begin{minipage}{0.7\linewidth}
%	\begin{center}
%	\includegraphics[width=0.8\linewidth]{figures/graphs/community_dataset_double_size.png}
%	\end{center}
%	\vspace{3mm}
%\end{minipage}
%\hfill
%\begin{minipage}{0.6\linewidth}
%	\setlength\fheight{3.2cm}
%	\setlength\fwidth{1.0\linewidth}
%	\hspace*{-3mm}
%	\input{figures/community_detection/ChebNet_CayleyNet_val_test_comparison_constant_signals_high_var_order_long_run_cap_size2.tex}	
%\\
%\hspace*{-3mm}
%	\input{figures/community_detection/Chebyshev_filter_responses_constant_signal_min_var_cheb_coords_all_range.tex}
%\\
%\hspace*{-3mm}
%	\input{figures/community_detection/CayleyNet_filter_responses_constant_signal_min_var_angles_all_range.tex}
%\end{minipage}
%\vspace{0mm}
%
%\vspace{0mm}
%\caption{Top: synthetic 15-communities graph. Second to the top: community detection accuracy of ChebNet and CayleyNet. Bottom two: normalized responses of four different filters learned by ChebNet (top) and CayleyNet (bottom), { each response in a different color}. Grey vertical lines represent the frequencies of the normalized Laplacian ($\tilde{\lambda} = 2\lambda_n^{-1} \lambda - 1$ for ChebNet and $C(\lambda) = (h\lambda - i)/(h\lambda + i)$ unrolled to a real line for CayleyNet). Note how thanks to spectral zoom property Cayley filters can focus on the band of low frequencies (dark grey lines) containing most of the information about communities.\vspace{-3mm}}
%\label{fig:perf-multiple-orders}
%\end{figure}

\subsection{Complexity}
\begin{figure}[!ht]
\centering 
\begin{minipage}{0.7\linewidth}
\setlength\fheight{1.6cm}
\setlength\fwidth{0.7\linewidth}
% This file was created by matplotlib2tikz v0.6.6.
\begin{tikzpicture}%[font=\fontsize{6}{8} \selectfont]

\definecolor{color1}{rgb}{1,0.498039215686275,0.0549019607843137}
\definecolor{color0}{rgb}{0.12156862745098,0.466666666666667,0.705882352941177}
\definecolor{color3}{rgb}{0.83921568627451,0.152941176470588,0.156862745098039}
\definecolor{color2}{rgb}{0.172549019607843,0.627450980392157,0.172549019607843}
\definecolor{color5}{rgb}{0.549019607843137,0.337254901960784,0.294117647058824}
\definecolor{color4}{rgb}{0.580392156862745,0.403921568627451,0.741176470588235}

\begin{axis}[
xlabel={Order $r$},
ylabel={Accuracy \%},
xmin=-0.3, xmax=6.3,
ymin=27.7668884323383, ymax=102.753363490908,
xtick={0,1,2,3,4,5,6},
xticklabels={1,3,5,7,9,11,13},
tick align=outside,
tick pos=left,
x grid style={lightgray!92.026143790849673!black},
y grid style={lightgray!92.026143790849673!black},
legend pos={south east},
legend cell align={left},
legend style={font=\fontsize{6}{8} \selectfont}
]
%\legend{CayleyNet\textsubscript{inv}, CayleyNet\textsubscript{1 iter}, CayleyNet\textsubscript{5 iter}, CayleyNet\textsubscript{9 iter}, CayleyNet\textsubscript{13 iter}, ChebNet}
\path [draw=color0, semithick] (axis cs:0,89.6277771629404)
--(axis cs:0,90.6579345069815);

\path [draw=color0, semithick] (axis cs:1,95.7759138587851)
--(axis cs:1,96.6812269493204);

\path [draw=color0, semithick] (axis cs:2,96.6688204729781)
--(axis cs:2,97.8454631268754);

\path [draw=color0, semithick] (axis cs:3,97.3625564601836)
--(axis cs:3,98.7517277500215);

\path [draw=color0, semithick] (axis cs:4,98.0490608308123)
--(axis cs:4,98.8652239897444);

\path [draw=color0, semithick] (axis cs:5,98.400952610021)
--(axis cs:5,99.256188808436);

\path [draw=color0, semithick] (axis cs:6,98.9408261490946)
--(axis cs:6,99.3448873518819);

\path [draw=color1, semithick] (axis cs:0,39.096495554597)
--(axis cs:0,43.9320775292805);

\path [draw=color1, semithick] (axis cs:1,68.2417559662256)
--(axis cs:1,74.7296714934912);

\path [draw=color1, semithick] (axis cs:2,80.4618809974169)
--(axis cs:2,85.2809747612501);

\path [draw=color1, semithick] (axis cs:3,86.4510145103728)
--(axis cs:3,90.3204130256379);

\path [draw=color1, semithick] (axis cs:4,88.6857134841777)
--(axis cs:4,91.7999992538594);

\path [draw=color1, semithick] (axis cs:5,91.5142847756774)
--(axis cs:5,92.9714265890688);

\path [draw=color1, semithick] (axis cs:6,92.1696707028157)
--(axis cs:6,94.030326855778);

\path [draw=color2, semithick] (axis cs:0,42.6289647102927)
--(axis cs:0,50.9424643372919);

\path [draw=color2, semithick] (axis cs:1,83.666802769292)
--(axis cs:1,88.3046240800732);

\path [draw=color2, semithick] (axis cs:2,87.2051096364615)
--(axis cs:2,90.3948893717173);

\path [draw=color2, semithick] (axis cs:3,89.8217175965372)
--(axis cs:3,92.8068506712851);

\path [draw=color2, semithick] (axis cs:4,92.8460780001049)
--(axis cs:4,95.4396333646413);

\path [draw=color2, semithick] (axis cs:5,95.8830612555224)
--(axis cs:5,97.1169373711866);

\path [draw=color2, semithick] (axis cs:6,96.1750062020614)
--(axis cs:6,97.9964209524797);

\path [draw=color3, semithick] (axis cs:0,55.5618578617072)
--(axis cs:0,65.1238567455316);

\path [draw=color3, semithick] (axis cs:1,87.066095606308)
--(axis cs:1,90.2481878409576);

\path [draw=color3, semithick] (axis cs:2,90.086748457863)
--(axis cs:2,94.0275353708723);

\path [draw=color3, semithick] (axis cs:3,92.4713695191035)
--(axis cs:3,95.9572004271855);

\path [draw=color3, semithick] (axis cs:4,95.8469893270208)
--(axis cs:4,97.867295340948);

\path [draw=color3, semithick] (axis cs:5,97.0551522748278)
--(axis cs:5,98.2877043230726);

\path [draw=color3, semithick] (axis cs:6,97.0976237860775)
--(axis cs:6,98.0452309807682);

\path [draw=color4, semithick] (axis cs:0,59.9117330096617)
--(axis cs:0,66.6311238171206);

\path [draw=color4, semithick] (axis cs:1,89.2891245299851)
--(axis cs:1,92.3965867584671);

\path [draw=color4, semithick] (axis cs:2,93.1392354307804)
--(axis cs:2,94.9464773072567);

\path [draw=color4, semithick] (axis cs:3,95.9968367529131)
--(axis cs:3,96.7745890283369);

\path [draw=color4, semithick] (axis cs:4,96.5726194898653)
--(axis cs:4,97.9416648729276);

\path [draw=color4, semithick] (axis cs:5,96.8168414594297)
--(axis cs:5,98.011728944623);

\path [draw=color4, semithick] (axis cs:6,97.7769199142982)
--(axis cs:6,98.3659353103112);

\path [draw=color5, semithick] (axis cs:0,31.1753645713642)
--(axis cs:0,33.453208121507);

\path [draw=color5, semithick] (axis cs:1,35.9405364529497)
--(axis cs:1,39.4594647486799);

\path [draw=color5, semithick] (axis cs:2,61.051043615415)
--(axis cs:2,65.2060997103907);

\path [draw=color5, semithick] (axis cs:3,72.873786750862)
--(axis cs:3,75.5547850646288);

\path [draw=color5, semithick] (axis cs:4,78.9462207987082)
--(axis cs:4,82.3394911000955);

\path [draw=color5, semithick] (axis cs:5,84.2248347823174)
--(axis cs:5,86.6323060257881);

\path [draw=color5, semithick] (axis cs:6,85.6490499742799)
--(axis cs:6,88.1795179120726);

\addplot [semithick, color0, mark=-, mark size=5, mark options={solid}, only marks]
table {%
0 89.6277771629404
1 95.7759138587851
2 96.6688204729781
3 97.3625564601836
4 98.0490608308123
5 98.400952610021
6 98.9408261490946
};
\addplot [semithick, color0, mark=-, mark size=5, mark options={solid}, only marks]
table {%
0 90.6579345069815
1 96.6812269493204
2 97.8454631268754
3 98.7517277500215
4 98.8652239897444
5 99.256188808436
6 99.3448873518819
};
\addplot [semithick, color1, mark=-, mark size=5, mark options={solid}, only marks]
table {%
0 39.096495554597
1 68.2417559662256
2 80.4618809974169
3 86.4510145103728
4 88.6857134841777
5 91.5142847756774
6 92.1696707028157
};
\addplot [semithick, color1, mark=-, mark size=5, mark options={solid}, only marks]
table {%
0 43.9320775292805
1 74.7296714934912
2 85.2809747612501
3 90.3204130256379
4 91.7999992538594
5 92.9714265890688
6 94.030326855778
};
\addplot [semithick, color2, mark=-, mark size=5, mark options={solid}, only marks]
table {%
0 42.6289647102927
1 83.666802769292
2 87.2051096364615
3 89.8217175965372
4 92.8460780001049
5 95.8830612555224
6 96.1750062020614
};
\addplot [semithick, color2, mark=-, mark size=5, mark options={solid}, only marks]
table {%
0 50.9424643372919
1 88.3046240800732
2 90.3948893717173
3 92.8068506712851
4 95.4396333646413
5 97.1169373711866
6 97.9964209524797
};
\addplot [semithick, color3, mark=-, mark size=5, mark options={solid}, only marks]
table {%
0 55.5618578617072
1 87.066095606308
2 90.086748457863
3 92.4713695191035
4 95.8469893270208
5 97.0551522748278
6 97.0976237860775
};
\addplot [semithick, color3, mark=-, mark size=5, mark options={solid}, only marks]
table {%
0 65.1238567455316
1 90.2481878409576
2 94.0275353708723
3 95.9572004271855
4 97.867295340948
5 98.2877043230726
6 98.0452309807682
};
\addplot [semithick, color4, mark=-, mark size=5, mark options={solid}, only marks]
table {%
0 59.9117330096617
1 89.2891245299851
2 93.1392354307804
3 95.9968367529131
4 96.5726194898653
5 96.8168414594297
6 97.7769199142982
};
\addplot [semithick, color4, mark=-, mark size=5, mark options={solid}, only marks]
table {%
0 66.6311238171206
1 92.3965867584671
2 94.9464773072567
3 96.7745890283369
4 97.9416648729276
5 98.011728944623
6 98.3659353103112
};
\addplot [semithick, color5, mark=-, mark size=5, mark options={solid}, only marks]
table {%
0 31.1753645713642
1 35.9405364529497
2 61.051043615415
3 72.873786750862
4 78.9462207987082
5 84.2248347823174
6 85.6490499742799
};
\addplot [semithick, color5, mark=-, mark size=5, mark options={solid}, only marks]
table {%
0 33.453208121507
1 39.4594647486799
2 65.2060997103907
3 75.5547850646288
4 82.3394911000955
5 86.6323060257881
6 88.1795179120726
};
\addplot [semithick, color0, dashed, mark=*, mark size=1, mark options={solid}]
table {%
0 90.142855834961
1 96.2285704040527
2 97.2571417999268
3 98.0571421051025
4 98.4571424102783
5 98.8285707092285
6 99.1428567504883
};
\addplot [semithick, color1, mark=*, mark size=1, mark options={solid}]
table {%
0 41.5142865419388
1 71.4857137298584
2 82.8714278793335
3 88.3857137680054
4 90.2428563690185
5 92.2428556823731
6 93.0999987792969
};
\addplot [semithick, color2, mark=*, mark size=1, mark options={solid}]
table {%
0 46.7857145237923
1 85.9857134246826
2 88.7999995040894
3 91.3142841339111
4 94.1428556823731
5 96.4999993133545
6 97.0857135772705
};
\addplot [semithick, color3, mark=*, mark size=1, mark options={solid}]
table {%
0 60.3428573036194
1 88.6571417236328
2 92.0571419143677
3 94.2142849731445
4 96.8571423339844
5 97.6714282989502
6 97.5714273834228
};
\addplot [semithick, color4, mark=*, mark size=1, mark options={solid}]
table {%
0 63.2714284133911
1 90.8428556442261
2 94.0428563690186
3 96.385712890625
4 97.2571421813965
5 97.4142852020264
6 98.0714276123047
};
\addplot [semithick, color5, dotted, mark=*, mark size=1, mark options={solid}]
table {%
0 32.3142863464356
1 37.7000006008148
2 63.1285716629028
3 74.2142859077454
4 80.6428559494018
5 85.4285704040527
6 86.9142839431763
};
\node[fill=white, font=\fontsize{8}{8}\selectfont, inner sep=0.1,outer sep=0] at (280,580) {1};
\node[fill=white, font=\fontsize{8}{8}\selectfont, inner sep=0.1,outer sep=0] at (280,620) {5};
\node[fill=white, font=\fontsize{8}{8}\selectfont, inner sep=0.1,outer sep=0] at (280,652) {9};
\node[fill=white, font=\fontsize{8}{8}\selectfont, inner sep=0.1,outer sep=0] at (280,680) {13};
%\node[fill=white, font=\fontsize{6}{8}\selectfont, inner sep=0.4,outer sep=0] at (6,83) {5};
%\node[fill=white, font=\fontsize{6}{8}\selectfont, inner sep=0.1,outer sep=0] at (6,86) {9};
%\node[fill=white, font=\fontsize{6}{8}\selectfont, inner sep=0.05,outer sep=0] at (6,88) {13};
\end{axis}
\end{tikzpicture}
\end{minipage}
\vspace{-1mm}
\caption{Community detection test accuracy as function of filter order $r$. Shown are exact matrix inversion (dashed) and approximate Jacobi with different number of iterations (colored). For reference, ChebNet is shown (dotted). \vspace{-1mm}}
\label{fig:perf-multiple-orders-multiple-jacobi-iter}
\end{figure}
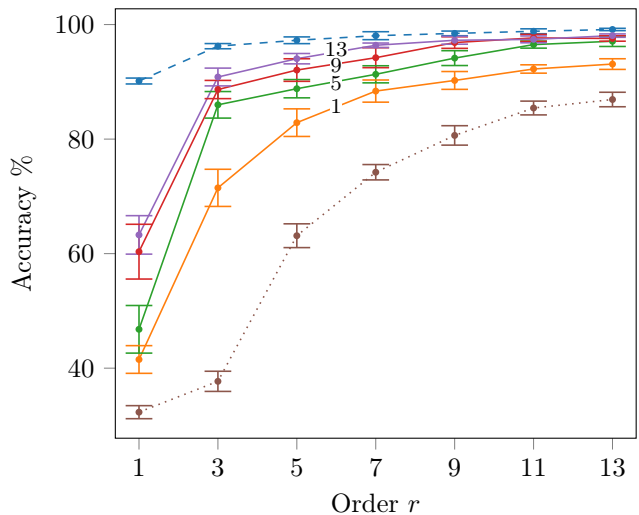

We experimentally validated the computational complexity of our model applying filters of different order $r$ to synthetic 15-community graphs of different size $n$ using exact matrix inversion and approximation with different number of Jacobi iterations (Figure \ref{fig:test-training-times-community-dataset} in Appendix C). { All times have been computed running 30 times the considered models and averaging the final results}. As expected, approximate inversion guarantees $\mathcal{O}(n)$ complexity. 
We further conclude that typically very few Jacobi iterations are required (Figure \ref{fig:perf-multiple-orders-multiple-jacobi-iter} shows that our model with just one Jacobi iteration outperforms ChebNet for low-order filters on the community detection problem). 

\subsection{MNIST}
Following \cite{defferrard2016convolutional,monti2016geometric}, \textcolor{black}{for a toy example,} we approached the classical MNIST digits classification as a learning problem on graphs. Each pixel of an image is a vertex of a graph (regular grid with 8-neighbor connectivity), and pixel color is a signal on the graph. 
We used a graph CNN architecture with two spectral convolutional layers based on Chebyshev and Cayley filters (producing 32 and 64 output features, respectively), interleaved with pooling layers performing 4-times graph coarsening using the Graclus algorithm \cite{dhillon2007weighted}, and finally a fully-connected layer (this architecture replicates the classical LeNet5, \cite{lecun1998gradient}, architecture, which is shown for comparison). { SGD+Momentum with learning rate equal to 0.02, momentum $m=0.9$, dropout probability $p=0.5$ and weight decay coefficient $\gamma = 5 \cdot 10^{-4}$ have been applied as described in \cite{defferrard2016convolutional}}.
MNIST classification results are reported in Table \ref{tab:MNIST}. CayleyNet {(11 Jacobi iterations)} achieves the same (near perfect) accuracy as ChebNet with filters of lower order ($r=12$ vs $25$).  
Examples of filters learned by ChebNet and CayleyNet are shown in Figure~\ref{fig:MNIST-filters}. { 0.1776 +/- 0.06079 sec and 0.0268 +/- 0.00841 sec are respectively required by CayleyNet and ChebNet for analyzing a batch of 100 images at test time.}

\vspace{0mm}
\begin{table}[!ht]
\centering
\caption{Test accuracy obtained with different methods on the MNIST dataset.}
\label{tab:MNIST}
\setlength\tabcolsep{1.5pt}
\begin{tabular}{@{}lccc@{}}
\toprule
Model     & Order  & Accuracy & \#Params  \\ \midrule
LeNet5    & - &99.33\% & 1.66M      \\
ChebNet   & 25 &99.14\%    & 1.66M  \\
CayleyNet & 12 & 99.18\%   & 1.66M        \\ \bottomrule
\end{tabular}
\end{table}

\subsection{Citation network} 
Next, we address the problem of vertex classification on graphs using the popular CORA citation graph \cite{sen2008collective}. 
Each of the 2,708 vertices of the CORA graph represents a scientific paper, and an undirected unweighted edge represents a citation
(5,429 edges in total). For each vertex, a 1,433-dimensional binary feature vector representing the content of the paper is given. The task is to classify each vertex into one of the 7 groundtruth classes, of labels. { In the semi-supervised problem (transductive learning), the features of all vertices are known, but labels are given just for a subset of the nodes. The task is to learn a mapping, that takes the features at the nodes as inputs, and gives the labels as outputs. The mapping is trained by minimizing the label error at the nodes with known labels. After training, the the mapping is tested over the nodes in which the labels were unknown during training.}

 To present a deep comparison with recent state-of-the-art architectures, we analyze the performance of our model in two different settings: the classic semi-supervised problem presented in  \cite{welling2016,monti2016geometric,velickovic2018} with 140 training samples, 500 validation samples and 1,000 test samples and a relaxed version of this that exploits 1,708 vertices for training, 500 for validation and 500 for testing. We opted for a larger amount of training samples in our second experiment, in order to provide an estimate of the quality of CayleyNet in a situation which is less prone to overfitting. This provides a better overview of the goodness of the considered construction, since richer filters are less likely to produce lower performance, as opposed to the typical behavior when the available data is scarce. Cayley operators with matrix inversion have been considered in both settings for our solution. DCNN\cite{atwood2016search}, GCN\cite{welling2016}, MoNet\cite{monti2016geometric} and GAT\cite{velickovic2018} have been used as terms of comparison.

 On the standard split, we train CayleyNet by realizing filters as linear combinations of neighborhood descriptors obtained by applying Cayley filters on  the signal of features. 
Two versions of CayleyNet have been implemented for this setting: a lightweight one exploiting two convolutional layers with 16 and 7 output features and a heavier one requiring 64 and 7 output features. This provides a valuable term of comparison with both the solutions presented in  \cite{welling2016,monti2016geometric} and \cite{velickovic2018} as same number of parameters is respectively required by our implementations. Normalized Laplacian has been used as reference. Adam with learning rate equal to $5 \cdot 10^{-3}$, dropout probability $p=0.6$ and weight decay coefficient $\gamma=5 \cdot 10^{-4}$ have been used for training. Table \ref{tab:cora-standard-split} presents the results we obtained with our solution, average performance over 50 runs are reported to guarantee accurate estimates. Performance of GCN, MoNet and GAT have been obtained from the respective papers, DCNN has been trained with 1 diffusion layer and 1 hop to guarantee same number of parameters. Our lighter version of CayleyNet outperforms DCNN, GCN and MoNet, while being defeated only by the recent GAT (which however exploits an attention mechanism for better discriminating relevant neighbors). Our heavier CayleyNet shows a significant drop in performance, likely because of overfitting on the small training set. 

 \begin{table}[!ht]
\begin{minipage}[t]{0.53\linewidth}
\centering
      \caption{Test accuracy of different methods on the standard split of the CORA dataset.}
      \label{tab:cora-standard-split}
\setlength\tabcolsep{1.5pt}
     \begin{tabular}{@{}lcc@{}}
\toprule
Method    & Accuracy &  \#Params\\ \midrule
DCNN \cite{atwood2016search} & 72.3 $\pm$ 0.8 \%  & 23K  \\
{\bf CayleyNet}\textsubscript{64 features} & {\bf 81.0 $\pm$ 0.5 \%}   & 92K  \\
GCN \cite{welling2016}   &  81.6 $\pm$ 0.4 \%   & 23K    \\
MoNet  \cite{monti2016geometric}   &   81.7 $\pm$ 0.5 \%    & 23K         \\ 
{\bf CayleyNet}\textsubscript{16 features} & {\bf 81.9 $\pm$ 0.7 \%}   & 23K    \\ 
GAT  \cite{velickovic2018}   &   83.0 $\pm$ 0.7 \%    & 92K         \\ 
\bottomrule
\end{tabular}
\end{minipage}
\hfill
\begin{minipage}[t]{0.43\linewidth}
\centering
      \caption{Test accuracy of different methods on the extended split of the CORA dataset.}
      \label{tab:cora}
\setlength\tabcolsep{1.5pt}
     \begin{tabular}{@{}lcc@{}}
\toprule
Method    & Accuracy &  \#Params\\ \midrule
DCNN  \cite{atwood2016search}   &   86.01 $\pm$ 0.24 \%    & 47K         \\ %& 86.60\%    & \textcolor{blue}{34K}          \\
GCN \cite{welling2016}   &  86.64 $\pm$ 0.55   \% & 47K    \\% & 87.17\%   & \textcolor{blue}{23K}    \\
ChebNet \cite{defferrard2016convolutional}  & 87.07 $\pm$ 0.72 \% & 46K      \\%& 87.12\% & \textcolor{blue}{46K}       \\
{\bf CayleyNet} & {\bf 88.09 $\pm$ 0.60 \%}   & 46K    \\ 
MoNet \cite{monti2016geometric} & 88.38 $\pm$ 0.46 \%   & 46K    \\ 
GAT \cite{velickovic2018} & 88.65 $\pm$ 0.58 \%   & 46K    \\ \bottomrule
\end{tabular}
\end{minipage}
\end{table}

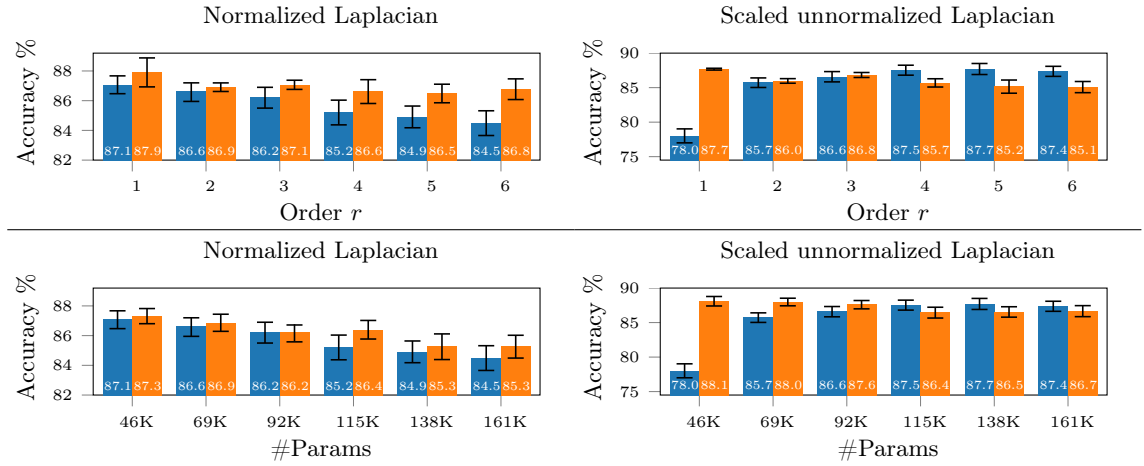
\begin{figure}[!ht]
\vspace{-3mm}
\centering 
 \begin{minipage}[t]{0.5\linewidth}
    \setlength\fheight{3.0cm}
\setlength\fwidth{1\linewidth}
% This file was created by matplotlib2tikz v0.6.7.
\begin{tikzpicture}[font=\fontsize{6}{8} \selectfont]

\definecolor{color1}{rgb}{1,0.498039215686275,0.0549019607843137}
\definecolor{color0}{rgb}{0.12156862745098,0.466666666666667,0.705882352941177}

\begin{axis}[
title={\small Normalized Laplacian},
xlabel={\small Order $r$},
ylabel={\small Accuracy \%},
xtick={1,2,3,4,5,6},
xmin=0.47, xmax=6.53,
ymin=82, ymax=89.2,
width=\fwidth,
height=\fheight,
ylabel shift = -5 pt,
xlabel shift = -2pt,
tick align=outside,
tick pos=left,
x grid style={lightgray!92.026143790849673!black},
y grid style={lightgray!92.026143790849673!black},
%legend style={draw=white!80.0!black},
%legend cell align={left},
%legend entries={{ChebNet},{CayleyNet}},
%legend style={at={(0.5,1.325)}, anchor=north,legend columns=-1, font=\small}
]
\addlegendimage{ybar,ybar legend,fill=color0,draw opacity=0};
\draw[fill=color0,draw opacity=0] (axis cs:0.6,0) rectangle (axis cs:1,87.071999);%87.120002746582);
\draw[fill=color0,draw opacity=0] (axis cs:1.6,0) rectangle (axis cs:2,86.607986);%86.89599609375);
\draw[fill=color0,draw opacity=0] (axis cs:2.6,0) rectangle (axis cs:3,86.216003);%85.2360076904297);
\draw[fill=color0,draw opacity=0] (axis cs:3.6,0) rectangle (axis cs:4,85.203995);%85.2360076904297);
\draw[fill=color0,draw opacity=0] (axis cs:4.6,0) rectangle (axis cs:5,84.907997);%85.2360076904297);
\draw[fill=color0,draw opacity=0] (axis cs:5.6,0) rectangle (axis cs:6,84.487999);%85.2360076904297);
\addlegendimage{ybar,ybar legend,fill=color1,draw opacity=0};
\draw[fill=color1,draw opacity=0] (axis cs:1,0) rectangle (axis cs:1.4,87.9039916992188);
\draw[fill=color1,draw opacity=0] (axis cs:2,0) rectangle (axis cs:2.4,86.9079971313477);
\draw[fill=color1,draw opacity=0] (axis cs:3,0) rectangle (axis cs:3.4,87.0680084228516);
\draw[fill=color1,draw opacity=0] (axis cs:4,0) rectangle (axis cs:4.4,86.612);
\draw[fill=color1,draw opacity=0] (axis cs:5,0) rectangle (axis cs:5.4,86.484016);
\draw[fill=color1,draw opacity=0] (axis cs:6,0) rectangle (axis cs:6.4,86.772003);

% vertical bars std chebNet
\path [draw=black, semithick] (axis cs:0.8,86.47)%87.0220230296254)
--(axis cs:0.8,87.67);%87.2179824635386);

\path [draw=black, semithick] (axis cs:1.8,85.95)%86.7960768938065)
--(axis cs:1.8,87.2);%86.9959152936935);

\path [draw=black, semithick] (axis cs:2.8,85.5)%85.0880081802607)
--(axis cs:2.8,86.9);%85.3840072005987);

\path [draw=black, semithick] (axis cs:3.8,84.37213994)%87.0220230296254)
--(axis cs:3.8,86.03585006);%87.2179824635386);

\path [draw=black, semithick] (axis cs:4.8,84.17674063)%86.7960768938065)
--(axis cs:4.8,85.63925337);%86.9959152936935);

\path [draw=black, semithick] (axis cs:5.8,83.65550003)%85.0880081802607)
--(axis cs:5.8,85.32049797);%85.3840072005987);

% vertical bars std cayleyNet

\path [draw=black, semithick] (axis cs:1.2,86.9289089441299)
--(axis cs:1.2,88.8790744543076);

\path [draw=black, semithick] (axis cs:2.2,86.6189711093903)
--(axis cs:2.2,87.1970231533051);

\path [draw=black, semithick] (axis cs:3.2,86.7591770887375)
--(axis cs:3.2,87.3768397569656);
            
\path [draw=black, semithick] (axis cs:4.2,85.8106)
--(axis cs:4.2,87.4134);

\path [draw=black, semithick] (axis cs:5.2,85.8591)
--(axis cs:5.2,87.1090);

\path [draw=black, semithick] (axis cs:6.2,86.0757)
--(axis cs:6.2,87.4683);

%\addplot [semithick, black, mark=-, mark size=3, mark options={solid}, only marks]
%table {%
%0.8 87.0220230296254
%1.8 86.7960768938065
%2.8 85.0880081802607
%};
\addplot [semithick, black, mark=-, mark size=3, mark options={solid}, only marks]
table {%
0.8 86.47
1.8 85.95
2.8 85.5
3.8 84.37213994
4.8 84.17674063
5.8 83.65550003
};
%\addplot [semithick, black, mark=-, mark size=3, mark options={solid}, only marks]
%table {%
%0.8 87.2179824635386
%1.8 86.9959152936935
%2.8 85.3840072005987
%};
\addplot [semithick, black, mark=-, mark size=3, mark options={solid}, only marks]
table {%
0.8 87.67
1.8 87.2
2.8 86.9
3.8 86.03585006
4.8 85.63925337
5.8 85.32049797
}; 
\addplot [semithick, black, mark=-, mark size=3, mark options={solid}, only marks]
table {%
1.2 86.9289089441299
2.2 86.6189711093903
3.2 86.7591770887375
4.2 85.8106
5.2 85.8591
6.2 86.0757
};
\addplot [semithick, black, mark=-, mark size=3, mark options={solid}, only marks]
table {%
1.2 88.8790744543076
2.2 87.1970231533051
3.2 87.3768397569656
4.2 87.4134
5.2 87.1090
6.2 87.4683
};

\node at (axis cs:0.76,81.85)[%84.05)[
  scale=0.79,
  anchor=south,
  text=white,
  rotate=0.0
]{ 87.1};
\node at (axis cs:1.76,81.85)[%84.05)[
  scale=0.79,
  anchor=south,
  text=white,
  rotate=0.0
]{ 86.6};%86.9};
\node at (axis cs:2.76,81.85)[%84.05)[
  scale=0.79,
  anchor=south,
  text=white,
  rotate=0.0
]{ 86.2};%85.2};
\node at (axis cs:3.76,81.85)[%84.05)[
  scale=0.79,
  anchor=south,
  text=white,
  rotate=0.0
]{ 85.2};%85.2};
\node at (axis cs:4.76,81.85)[%84.05)[
  scale=0.79,
  anchor=south,
  text=white,
  rotate=0.0
]{ 84.9};%85.2};
\node at (axis cs:5.76,81.85)[%84.05)[
  scale=0.79,
  anchor=south,
  text=white,
  rotate=0.0
]{ 84.5};%85.2};

\node at (axis cs:1.16,81.85)[%84.05)[
  scale=0.79,
  anchor=south,
  text=white,
  rotate=0.0
]{ 87.9};
\node at (axis cs:2.16,81.85)[%84.05)[
  scale=0.79,
  anchor=south,
  text=white,
  rotate=0.0
]{ 86.9};
\node at (axis cs:3.16,81.85)[%84.05)[
  scale=0.79,
  anchor=south,
  text=white,
  rotate=0.0
]{ 87.1};
\node at (axis cs:4.16,81.85)[%84.05)[
  scale=0.79,
  anchor=south,
  text=white,
  rotate=0.0
]{ 86.6};
\node at (axis cs:5.16,81.85)[%84.05)[
  scale=0.79,
  anchor=south,
  text=white,
  rotate=0.0
]{ 86.5};
\node at (axis cs:6.16,81.85)[%84.05)[
  scale=0.79,
  anchor=south,
  text=white,
  rotate=0.0
]{ 86.8};

\end{axis}

\end{tikzpicture}
\hrule
\hfill
\end{minipage}\hfill
\begin{minipage}[t]{0.5\linewidth}
    \setlength\fheight{3.0cm}
\setlength\fwidth{1\linewidth}
% This file was created by matplotlib2tikz v0.6.7.
\begin{tikzpicture}[font=\fontsize{6}{8} \selectfont]

\definecolor{color1}{rgb}{1,0.498039215686275,0.0549019607843137}
\definecolor{color0}{rgb}{0.12156862745098,0.466666666666667,0.705882352941177}

\begin{axis}[
title={\small Scaled unnormalized Laplacian},
xlabel={\small Order $r$},
ylabel={\small Accuracy \%},
xtick={1,2,3,4,5,6},
xmin=0.47, xmax=6.53,
ymin=74.5, ymax=90,
width=\fwidth,
height=\fheight,
ylabel shift = -5 pt,
xlabel shift = -2pt,
tick align=outside,
tick pos=left,
x grid style={lightgray!92.026143790849673!black},
y grid style={lightgray!92.026143790849673!black},
%legend style={draw=white!80.0!black},
%legend cell align={left},
%legend entries={{ChebNet},{CayleyNet}},
%legend style={at={(0.5,1.325)}, anchor=north,legend columns=-1, font=\small}
]
\addlegendimage{ybar,ybar legend,fill=color0,draw opacity=0};
\draw[fill=color0,draw opacity=0] (axis cs:0.6,0) rectangle (axis cs:1,78.011993);%87.120002746582);
\draw[fill=color0,draw opacity=0] (axis cs:1.6,0) rectangle (axis cs:2,85.715996);%86.89599609375);
\draw[fill=color0,draw opacity=0] (axis cs:2.6,0) rectangle (axis cs:3,86.580002);%86.216003
\draw[fill=color0,draw opacity=0] (axis cs:3.6,0) rectangle (axis cs:4,87.528008);%85.2360076904297);
\draw[fill=color0,draw opacity=0] (axis cs:4.6,0) rectangle (axis cs:5,87.700012);%85.2360076904297);
\draw[fill=color0,draw opacity=0] (axis cs:5.6,0) rectangle (axis cs:6,87.360008);%85.2360076904297);
\addlegendimage{ybar,ybar legend,fill=color1,draw opacity=0};
\draw[fill=color1,draw opacity=0] (axis cs:1,0) rectangle (axis cs:1.4,87.676003);
\draw[fill=color1,draw opacity=0] (axis cs:2,0) rectangle (axis cs:2.4,85.972);
\draw[fill=color1,draw opacity=0] (axis cs:3,0) rectangle (axis cs:3.4,86.832001);
\draw[fill=color1,draw opacity=0] (axis cs:4,0) rectangle (axis cs:4.4,85.688011);
\draw[fill=color1,draw opacity=0] (axis cs:5,0) rectangle (axis cs:5.4,85.14801);
\draw[fill=color1,draw opacity=0] (axis cs:6,0) rectangle (axis cs:6.4,85.080002);

\path [draw=black, semithick] (axis cs:0.8,76.9997)%87.0220230296254)
--(axis cs:0.8,79.0242);%87.2179824635386);

\path [draw=black, semithick] (axis cs:1.8,85.0231)%86.7960768938065)
--(axis cs:1.8,86.4089);%86.9959152936935);

\path [draw=black, semithick] (axis cs:2.8,85.8346)%85.0880081802607)
--(axis cs:2.8,87.3254);%85.3840072005987);

\path [draw=black, semithick] (axis cs:3.8,86.8061)%87.0220230296254)
--(axis cs:3.8,88.2500);%87.2179824635386);

\path [draw=black, semithick] (axis cs:4.8,86.9008)%86.7960768938065)
--(axis cs:4.8,88.4993);%86.9959152936935);

\path [draw=black, semithick] (axis cs:5.8,86.6312)%85.0880081802607)
--(axis cs:5.8,88.0888);%85.3840072005987);     

\path [draw=black, semithick] (axis cs:1.2,87.5440029352903)
--(axis cs:1.2,87.8080020695925);

\path [draw=black, semithick] (axis cs:2.2,85.6397120058537)
--(axis cs:2.2,86.304288238287);

\path [draw=black, semithick] (axis cs:3.2,86.4712010622025)
--(axis cs:3.2,87.1928004026413);
      
\path [draw=black, semithick] (axis cs:4.2,85.0891)
--(axis cs:4.2,86.2869);

\path [draw=black, semithick] (axis cs:5.2,84.1932)
--(axis cs:5.2,86.1028);

\path [draw=black, semithick] (axis cs:6.2,84.2711)
--(axis cs:6.2,85.8890);

%\addplot [semithick, black, mark=-, mark size=3, mark options={solid}, only marks]
%table {%
%0.8 87.0220230296254
%1.8 86.7960768938065
%2.8 85.0880081802607
%};
\addplot [semithick, black, mark=-, mark size=3, mark options={solid}, only marks]
table {%
0.8 76.9997
1.8 85.0231
2.8 85.8346
3.8 86.8061
4.8 86.9008
5.8 86.6312
};
%\addplot [semithick, black, mark=-, mark size=3, mark options={solid}, only marks]
%table {%
%0.8 87.2179824635386
%1.8 86.9959152936935
%2.8 85.3840072005987
%};
\addplot [semithick, black, mark=-, mark size=3, mark options={solid}, only marks]
table {%
0.8 79.0242
1.8 86.4089
2.8 87.3254 
3.8 88.2500
4.8 88.4993
5.8 88.0888
};             
\addplot [semithick, black, mark=-, mark size=3, mark options={solid}, only marks]
table {%
1.2 87.5440029352903
2.2 85.6397120058537
3.2 86.4712010622025
4.2 85.0891
5.2 84.1932
6.2 84.2711
};
\addplot [semithick, black, mark=-, mark size=3, mark options={solid}, only marks]
table {%
1.2 87.8080020695925
2.2 86.304288238287
3.2 87.1928004026413
4.2 86.2869
5.2 86.1028
6.2 85.8890
};

\node at (axis cs:0.76,74.35)[%84.05)[
  scale=0.79,
  anchor=south,
  text=white,
  rotate=0.0
]{ 78.0};
\node at (axis cs:1.76,74.35)[%84.05)[
  scale=0.79,
  anchor=south,
  text=white,
  rotate=0.0
]{ 85.7};%86.9};
\node at (axis cs:2.76,74.35)[%84.05)[
  scale=0.79,
  anchor=south,
  text=white,
  rotate=0.0
]{ 86.6};%85.2};
\node at (axis cs:3.76,74.35)[%84.05)[
  scale=0.79,
  anchor=south,
  text=white,
  rotate=0.0
]{ 87.5};%85.2};
\node at (axis cs:4.76,74.35)[%84.05)[
  scale=0.79,
  anchor=south,
  text=white,
  rotate=0.0
]{ 87.7};%85.2};
\node at (axis cs:5.76,74.35)[%84.05)[
  scale=0.79,
  anchor=south,
  text=white,
  rotate=0.0
]{ 87.4};%85.2};
\node at (axis cs:1.16,74.35)[
  scale=0.79,
  anchor=south,
  text=white,
  rotate=0.0
]{ 87.7};
\node at (axis cs:2.16,74.35)[
  scale=0.79,
  anchor=south,
  text=white,
  rotate=0.0
]{ 86.0};
\node at (axis cs:3.16,74.35)[
  scale=0.79,
  anchor=south,
  text=white,
  rotate=0.0
]{ 86.8};
\node at (axis cs:4.16,74.35)[
  scale=0.79,
  anchor=south,
  text=white,
  rotate=0.0
]{ 85.7};
\node at (axis cs:5.16,74.35)[
  scale=0.79,
  anchor=south,
  text=white,
  rotate=0.0
]{ 85.2};
\node at (axis cs:6.16,74.35)[
  scale=0.79,
  anchor=south,
  text=white,
  rotate=0.0
]{ 85.1};
\end{axis}

\end{tikzpicture}
\hrule
\end{minipage}
 \begin{minipage}[t]{0.5\linewidth}
    \setlength\fheight{3.0cm}
\setlength\fwidth{1\linewidth}
% This file was created by matplotlib2tikz v0.6.7.
\begin{tikzpicture}[font=\fontsize{6}{8} \selectfont]

\definecolor{color1}{rgb}{1,0.498039215686275,0.0549019607843137}
\definecolor{color0}{rgb}{0.12156862745098,0.466666666666667,0.705882352941177}

\begin{axis}[
title={\small Normalized Laplacian},
xlabel={\small \#Params},
ylabel={\small Accuracy \%},
xtick={1,2,3,4,5,6},
xticklabels={46K, 69K, 92K, 115K, 138K, 161K},
xmin=0.47, xmax=6.53,
ymin=82, ymax=89.2,
width=\fwidth,
height=\fheight,
ylabel shift = -5 pt,
xlabel shift = -2pt,
tick align=outside,
tick pos=left,
x grid style={lightgray!92 s.026143790849673!black},
y grid style={lightgray!92.026143790849673!black},
%legend style={draw=white!80.0!black},
%legend cell align={left},
%legend entries={{ChebNet},{CayleyNet}},
%legend style={at={(0.5,1.325)}, anchor=north,legend columns=-1, font=\small}
]
\addlegendimage{ybar,ybar legend,fill=color0,draw opacity=0};
\draw[fill=color0,draw opacity=0] (axis cs:0.6,0) rectangle (axis cs:1,87.071999);%87.120002746582);
\draw[fill=color0,draw opacity=0] (axis cs:1.6,0) rectangle (axis cs:2,86.607986);%86.89599609375);
\draw[fill=color0,draw opacity=0] (axis cs:2.6,0) rectangle (axis cs:3,86.2);%86.216003
\draw[fill=color0,draw opacity=0] (axis cs:3.6,0) rectangle (axis cs:4,85.203995);%85.2360076904297);
\draw[fill=color0,draw opacity=0] (axis cs:4.6,0) rectangle (axis cs:5,84.907997);%85.2360076904297);
\draw[fill=color0,draw opacity=0] (axis cs:5.6,0) rectangle (axis cs:6,84.487999);%85.2360076904297);
\addlegendimage{ybar,ybar legend,fill=color1,draw opacity=0};
\draw[fill=color1,draw opacity=0] (axis cs:1,0) rectangle (axis cs:1.4,87.311989);
\draw[fill=color1,draw opacity=0] (axis cs:2,0) rectangle (axis cs:2.4,86.863991);
\draw[fill=color1,draw opacity=0] (axis cs:3,0) rectangle (axis cs:3.4,86.2); % 86.147995
\draw[fill=color1,draw opacity=0] (axis cs:4,0) rectangle (axis cs:4.4,86.395996);
\draw[fill=color1,draw opacity=0] (axis cs:5,0) rectangle (axis cs:5.4,85.251991);
\draw[fill=color1,draw opacity=0] (axis cs:6,0) rectangle (axis cs:6.4,85.255997);

% vertical bars std chebNet
\path [draw=black, semithick] (axis cs:0.8,86.47)%87.0220230296254)
--(axis cs:0.8,87.67);%87.2179824635386);

\path [draw=black, semithick] (axis cs:1.8,85.95)%86.7960768938065)
--(axis cs:1.8,87.2);%86.9959152936935);

\path [draw=black, semithick] (axis cs:2.8,85.5)%85.0880081802607)
--(axis cs:2.8,86.9);%85.3840072005987);

\path [draw=black, semithick] (axis cs:3.8,84.37213994)%87.0220230296254)
--(axis cs:3.8,86.03585006);%87.2179824635386);

\path [draw=black, semithick] (axis cs:4.8,84.17674063)%86.7960768938065)
--(axis cs:4.8,85.63925337);%86.9959152936935);

\path [draw=black, semithick] (axis cs:5.8,83.65550003)%85.0880081802607)
--(axis cs:5.8,85.32049797);%85.3840072005987);

% vertical bars std cayleyNet

\path [draw=black, semithick] (axis cs:1.2,86.8026)
--(axis cs:1.2,87.8214);

\path [draw=black, semithick] (axis cs:2.2,86.2879)
--(axis cs:2.2,87.4401);

\path [draw=black, semithick] (axis cs:3.2,85.6298)
--(axis cs:3.2,86.7662);

\path [draw=black, semithick] (axis cs:4.2,85.7705)
--(axis cs:4.2,87.0214);

\path [draw=black, semithick] (axis cs:5.2,84.3887)
--(axis cs:5.2,86.1153);

\path [draw=black, semithick] (axis cs:6.2,84.4886)
--(axis cs:6.2,86.0234);

%\addplot [semithick, black, mark=-, mark size=3, mark options={solid}, only marks]
%table {%
%0.8 87.0220230296254
%1.8 86.7960768938065
%2.8 85.0880081802607
%};
\addplot [semithick, black, mark=-, mark size=3, mark options={solid}, only marks]
table {%
0.8 86.47
1.8 85.95
2.8 85.5
3.8 84.37213994
4.8 84.17674063
5.8 83.65550003
};
%\addplot [semithick, black, mark=-, mark size=3, mark options={solid}, only marks]
%table {%
%0.8 87.2179824635386
%1.8 86.9959152936935
%2.8 85.3840072005987
%};
\addplot [semithick, black, mark=-, mark size=3, mark options={solid}, only marks]
table {%
0.8 87.67
1.8 87.2
2.8 86.9
3.8 86.03585006
4.8 85.63925337
5.8 85.32049797
};       
\addplot [semithick, black, mark=-, mark size=3, mark options={solid}, only marks]
table {%
1.2 86.8026
2.2 86.2879
3.2 85.5798
4.2 85.7705
5.2 84.3887
6.2 84.4886
};              
\addplot [semithick, black, mark=-, mark size=3, mark options={solid}, only marks]
table {%
1.2 87.8214
2.2 87.4401
3.2 86.7162
4.2 87.0214
5.2 86.1153
6.2 86.0234
};

\node at (axis cs:0.76,81.85)[%84.05)[
  scale=0.79,
  anchor=south,
  text=white,
  rotate=0.0
]{ 87.1};
\node at (axis cs:1.76,81.85)[%84.05)[
  scale=0.79,
  anchor=south,
  text=white,
  rotate=0.0
]{ 86.6};%86.9};
\node at (axis cs:2.76,81.85)[%84.05)[
  scale=0.79,
  anchor=south,
  text=white,
  rotate=0.0
]{ 86.2};%85.2};
\node at (axis cs:3.76,81.85)[%84.05)[
  scale=0.79,
  anchor=south,
  text=white,
  rotate=0.0
]{ 85.2};%85.2};
\node at (axis cs:4.76,81.85)[%84.05)[
  scale=0.79,
  anchor=south,
  text=white,
  rotate=0.0
]{ 84.9};%85.2};
\node at (axis cs:5.76,81.85)[%84.05)[
  scale=0.79,
  anchor=south,
  text=white,
  rotate=0.0
]{ 84.5};%85.2};
\node at (axis cs:1.16,81.85)[%84.05)[
  scale=0.79,
  anchor=south,
  text=white,
  rotate=0.0
]{87.3};
\node at (axis cs:2.16,81.85)[%84.05)[
  scale=0.79,
  anchor=south,
  text=white,
  rotate=0.0
]{86.9};
\node at (axis cs:3.16,81.85)[%84.05)[
  scale=0.79,
  anchor=south,
  text=white,
  rotate=0.0
]{ 86.2};
\node at (axis cs:4.16,81.85)[%84.05)[
  scale=0.79,
  anchor=south,
  text=white,
  rotate=0.0
]{86.4};
\node at (axis cs:5.16,81.85)[%84.05)[
  scale=0.79,
  anchor=south,
  text=white,
  rotate=0.0
]{85.3};
\node at (axis cs:6.16,81.85)[%84.05)[
  scale=0.79,
  anchor=south,
  text=white,
  rotate=0.0
]{85.3};
\end{axis}

\end{tikzpicture}
\hrule
\end{minipage}\hfill
 \begin{minipage}[t]{0.5\linewidth}
    \setlength\fheight{3.0cm}
\setlength\fwidth{1\linewidth}
% This file was created by matplotlib2tikz v0.6.7.
\begin{tikzpicture}[font=\fontsize{6}{8} \selectfont]

\definecolor{color1}{rgb}{1,0.498039215686275,0.0549019607843137}
\definecolor{color0}{rgb}{0.12156862745098,0.466666666666667,0.705882352941177}

\begin{axis}[
title={\small Scaled unnormalized Laplacian},
xlabel={\small \#Params},
ylabel={\small Accuracy \%},
xtick={1,2,3,4,5,6},
xticklabels={46K, 69K, 92K, 115K, 138K, 161K},
xmin=0.47, xmax=6.53,
ymin=74.5, ymax=90,
width=\fwidth,
height=\fheight,
ylabel shift = -5 pt,
xlabel shift = -2pt,
tick align=outside,
tick pos=left,
x grid style={lightgray!92 s.026143790849673!black},
y grid style={lightgray!92.026143790849673!black},
%legend style={draw=white!80.0!black},
%legend cell align={left},
%legend entries={{ChebNet},{CayleyNet}},
%legend style={at={(0.5,1.325)}, anchor=north,legend columns=-1, font=\small}
]

\addlegendimage{ybar,ybar legend,fill=color0,draw opacity=0};
\draw[fill=color0,draw opacity=0] (axis cs:0.6,0) rectangle (axis cs:1,78.011993);%87.120002746582);
\draw[fill=color0,draw opacity=0] (axis cs:1.6,0) rectangle (axis cs:2,85.715996);%86.89599609375);
\draw[fill=color0,draw opacity=0] (axis cs:2.6,0) rectangle (axis cs:3,86.580002);%86.216003
\draw[fill=color0,draw opacity=0] (axis cs:3.6,0) rectangle (axis cs:4,87.528008);%85.2360076904297);
\draw[fill=color0,draw opacity=0] (axis cs:4.6,0) rectangle (axis cs:5,87.700012);%85.2360076904297);
\draw[fill=color0,draw opacity=0] (axis cs:5.6,0) rectangle (axis cs:6,87.360008);%85.2360076904297);
\addlegendimage{ybar,ybar legend,fill=color1,draw opacity=0};
\draw[fill=color1,draw opacity=0] (axis cs:1,0) rectangle (axis cs:1.4,88.092003);
\draw[fill=color1,draw opacity=0] (axis cs:2,0) rectangle (axis cs:2.4,87.980003);
\draw[fill=color1,draw opacity=0] (axis cs:3,0) rectangle (axis cs:3.4, 87.59201); % 86.147995
\draw[fill=color1,draw opacity=0] (axis cs:4,0) rectangle (axis cs:4.4,86.440002);
\draw[fill=color1,draw opacity=0] (axis cs:5,0) rectangle (axis cs:5.4,86.540001);
\draw[fill=color1,draw opacity=0] (axis cs:6,0) rectangle (axis cs:6.4,86.656013);

% vertical bars std chebNet
\path [draw=black, semithick] (axis cs:0.8,76.9997)%87.0220230296254)
--(axis cs:0.8,79.0242);%87.2179824635386);

\path [draw=black, semithick] (axis cs:1.8,85.0231)%86.7960768938065)
--(axis cs:1.8,86.4089);%86.9959152936935);

\path [draw=black, semithick] (axis cs:2.8,85.8346)%85.0880081802607)
--(axis cs:2.8,87.3254);%85.3840072005987);

\path [draw=black, semithick] (axis cs:3.8,86.8061)%87.0220230296254)
--(axis cs:3.8,88.2500);%87.2179824635386);

\path [draw=black, semithick] (axis cs:4.8,86.9008)%86.7960768938065)
--(axis cs:4.8,88.4993);%86.9959152936935);

\path [draw=black, semithick] (axis cs:5.8,86.6312)%85.0880081802607)
--(axis cs:5.8,88.0888);%85.3840072005987);

% vertical bars std cayleyNet

\path [draw=black, semithick] (axis cs:1.2,87.4047)
--(axis cs:1.2,88.7793 );

\path [draw=black, semithick] (axis cs:2.2,87.4297)
--(axis cs:2.2,88.5303);

\path [draw=black, semithick] (axis cs:3.2,86.9881)
--(axis cs:3.2,88.1959);

\path [draw=black, semithick] (axis cs:4.2,85.6572)
--(axis cs:4.2,87.2228);

\path [draw=black, semithick] (axis cs:5.2,85.7861)
--(axis cs:5.2,87.2939);

\path [draw=black, semithick] (axis cs:6.2,85.8580)
--(axis cs:6.2,87.4541);

%\addplot [semithick, black, mark=-, mark size=3, mark options={solid}, only marks]
%table {%
%0.8 87.0220230296254
%1.8 86.7960768938065
%2.8 85.0880081802607
%};          
\addplot [semithick, black, mark=-, mark size=3, mark options={solid}, only marks]
table {%
0.8 76.9997
1.8 85.0231
2.8 85.8346
3.8 86.8061
4.8 86.9008
5.8 86.6312
};
%\addplot [semithick, black, mark=-, mark size=3, mark options={solid}, only marks]
%table {%
%0.8 87.2179824635386
%1.8 86.9959152936935
%2.8 85.3840072005987
%};
\addplot [semithick, black, mark=-, mark size=3, mark options={solid}, only marks]
table {%
0.8 79.0242
1.8 86.4089
2.8 87.3254 
3.8 88.2500
4.8 88.4993
5.8 88.0888
};                   
\addplot [semithick, black, mark=-, mark size=3, mark options={solid}, only marks]
table {%
1.2 87.4047
2.2 87.4297
3.2 86.9881
4.2 85.6572
5.2 85.7861
6.2 85.8580
};                      
\addplot [semithick, black, mark=-, mark size=3, mark options={solid}, only marks]
table {%
1.2 88.7793
2.2 88.5303
3.2 88.1959
4.2 87.2228
5.2 87.2939
6.2 87.4541
};

\node at (axis cs:0.76,74.35)[%84.05)[
  scale=0.79,
  anchor=south,
  text=white,
  rotate=0.0
]{ 78.0};
\node at (axis cs:1.76,74.35)[%84.05)[
  scale=0.79,
  anchor=south,
  text=white,
  rotate=0.0
]{ 85.7};%86.9};
\node at (axis cs:2.76,74.35)[%84.05)[
  scale=0.79,
  anchor=south,
  text=white,
  rotate=0.0
]{ 86.6};%85.2};
\node at (axis cs:3.76,74.35)[%84.05)[
  scale=0.79,
  anchor=south,
  text=white,
  rotate=0.0
]{ 87.5};%85.2};
\node at (axis cs:4.76,74.35)[%84.05)[
  scale=0.79,
  anchor=south,
  text=white,
  rotate=0.0
]{ 87.7};%85.2};
\node at (axis cs:5.76,74.35)[%84.05)[
  scale=0.79,
  anchor=south,
  text=white,
  rotate=0.0
]{ 87.4};%85.2};
\node at (axis cs:1.16,74.35)[%84.05)[
  scale=0.79,
  anchor=south,
  text=white,
  rotate=0.0
]{88.1};
\node at (axis cs:2.16,74.35)[%84.05)[
  scale=0.79,
  anchor=south,
  text=white,
  rotate=0.0
]{88.0};
\node at (axis cs:3.16,74.35)[%84.05)[
  scale=0.79,
  anchor=south,
  text=white,
  rotate=0.0
]{ 87.6};
\node at (axis cs:4.16,74.35)[%84.05)[
  scale=0.79,
  anchor=south,
  text=white,
  rotate=0.0
]{86.4};
\node at (axis cs:5.16,74.35)[%84.05)[
  scale=0.79,
  anchor=south,
  text=white,
  rotate=0.0
]{86.5};
\node at (axis cs:6.16,74.35)[%84.05)[
  scale=0.79,
  anchor=south,
  text=white,
  rotate=0.0
]{86.7};
\end{axis}

\end{tikzpicture}
\end{minipage}
\caption{ChebNet (blue) and CayleyNet (orange) test accuracies obtained on the CORA dataset for different polynomial orders. { Polynomials with complex coefficients (top two) and real coefficients (bottom two) have been exploited with CayleyNet in the two analysis. Orders 1 to 6 have been used in both comparisons. }}
\label{fig:CORA-perf} 
\vspace{-1mm}
\end{figure}

On our extended split, we analyze the behavior of CayleyNet and ChebNet for a variety of different polynomial orders. 
Two spectral convolutional layers with 16 and 7 outputs features have been used for implementing the two architectures.  { Adam with learning rate equal to $10^{-3}$, dropout probability $p=0.5$ and weight decay with coefficient $\gamma = 5 \cdot 10^{-4}$ have been used for training.} Figure \ref{fig:CORA-perf} presents the results of our analysis. Since ChebNet requires Laplacians with spectra bounded in $[-1,1]$, we consider both the normalized Laplacian (the two left figures), and the scaled unnormalized Laplacian $(2\boldsymbol{\Delta}/{\lambda_{max}}-\mathbf{I})$, where $\boldsymbol{\Delta}$ is the unnormalized Laplacian and $\lambda_{max}$ is its largest eigenvalue (the two right figures). For fair comparison, we fix the order of the filters (top figures) and the overall number of network parameters (bottom figures). In the bottom figures, the Cayley filters are restricted to even cosine polynomials by considering only real filter coefficients. The best CayleyNets consistently outperform the best ChebNets requiring at the same time less parameters (CayleyNet with order $r$ and complex coefficients requires a number of parameters equal to ChebNet with order $2r$).
To further complete our analysis, we present the performance obtained by DCNN, GCN, MoNet and GAT on our extended split (Table~\ref{tab:cora}). Two convolutional layers with order $r=1$, 1 head / 1 gaussian kernel, 16 and 7 outputs features have been used for GAT and MoNet\footnote{Filters have been realized as: $\mathbf{X}\mathbf{W_0} + \tilde{\mathbf{A}}\mathbf{X}\mathbf{W_1}$; with $\tilde{\mathbf{A}}$ the corresponding learned adjacency matrix. Attention has been computed for GAT only at second layer to ensure same number of parameters.}; 3 convolutional layers with 32 and 16 hidden features have been used for GCN; 2 diffusion layers with 10 hidden features and 2 diffusion hops for DCNN. GCN, MoNet and GAT have been trained with mean cross-entropy, dropout probability $p=0.5$ and weight decay with coefficient $\gamma = 5 \cdot 10^{-4}$, DCNN has been trained with hinge loss and no regularization (as reported in \cite{atwood2016search}). CayleyNet appears as the third best approach for solving the considered semi-supervised classification task, and outperforms other spectral CNN methods.

\subsection{Recommender system}

In our final experiment, we applied CayleyNet to recommendation system, formulated as  matrix completion problem on user and item graphs, \cite{monti2017geometric}. The task is, given a sparsely sampled matrix of scores assigned by users (columns) to items (rows), to fill in the missing scores. The similarities between users and items are given in the form of column and row graphs, respectively.  
\cite{monti2017geometric} approached this problem as learning with a Recurrent Graph CNN (RGCNN) architecture, using an extension of ChebNets to matrices defined on multiple graphs in order to extract spatial features from the score matrix; these features are then fed into an RNN producing a sequential estimation of the missing scores.  
Here, we repeated verbatim their experiment on the MovieLens dataset (\cite{miller2003movielens}), replacing Chebyshev filters with Cayley filters. { Following \cite{monti2017geometric}, to train our model we uniformly split the available training scores in two sets of equal dimension, 50\% of the provided scores (data scores) are used to initialize the input matrix while the remaining 50\% are used as training labels. SRGCNN is trained to reconstruct the missing labels from the few given data scores. At test time we initialize the input matrix only with the considered data scores to provide the network the same conditions it observed at training time.}
We used separable RGCNN architecture with two CayleyNets of order $r=4$ employing 15 Jacobi iterations. { Adam with learning rate equal to $10^{-3}$ and regularization coefficient $\gamma=10^{-10}$ have been used for training\footnote{Values obtained from MGCNN implementation available at \url{https://github.com/fmonti/mgcnn/}.}}
The results are reported in Table \ref{tab:results-movielens}. { To present a complete comparison we further extended the experiments reported in \cite{monti2017geometric} by  training sRGCNN with ChebNets of order 8, this provides an architecture with same number of parameters as the exploited CayleyNet (23K coefficients)}. Our version of sRGCNN outperforms all the competing methods, including the previous result with Chebyshev filters reported in \cite{monti2017geometric}. { sRGCNNs with Chebyshev polynomials of order 4 and 8 respectively require 0.0698 +/- 0.00275 sec and 0.0877 +/- 0.00362 sec at test time, sRGCNN with Cayley polynomials of order 4 and 15 jacobi iterations requires 0.165 +/- 0.00332 sec.}

\vspace{-1mm}
\begin{table}[!ht]
\centering
\caption{Performance (RMSE) of different matrix completion methods on the MovieLens dataset.}
\label{tab:results-movielens}
\setlength\tabcolsep{1pt}
\begin{tabular}{lc}
\toprule
Method & RMSE \\
 \midrule
MC \cite{candes2012exact} & 0.973\\
IMC \cite{jain2013provable,xu2013speedup} & 1.653\\
GMC  \cite{kalofolias2014matrix} & 0.996\\
%\hline
GRALS \cite{rao2015collaborative} & 0.945\\  %\midrule
sRGCNN$_{\rm Cheby, r=4}$ \cite{monti2017geometric}  & 0.929 \\
sRGCNN$_{\rm Cheby, r=8}$ \cite{monti2017geometric}  & 0.925 \\
{\bf sRGCNN$_{\mathbf{Cayley}}$} & {\bf 0.922} \\
\bottomrule
\end{tabular}
\hfill
\end{table}

\section{Conclusion}
In this paper, we introduced a new efficient spectral graph CNN architecture that scales linearly with the dimension of the input data. Our architecture is based on a new class of complex rational Cayley filters that are localized in space, can represent any smooth spectral transfer function, and are highly regular. The key property of our model is its ability to specialize in narrow frequency bands with a small number of filter parameters, while still preserving locality in the spatial domain.  
{
Experimental results on the MNIST, CORA and MovieLens datasets show the good performance of our construction wrt a variety of other approaches, and the superior performance with respect to other spectral CNN methods.}

\appendix

\section{Proof of Proposition 1}
First note the following classical result for the approximation of $\mathbf{A}\mathbf{x}=\mathbf{b}$ using the Jacobi method:
if the initial condition is $\mathbf{x}^{(0)}=\mathbf{0}$, then $(\mathbf{x}-\mathbf{x}^{(k)}) = \mathbf{J}^k\mathbf{x}$. In our case, note that if we start with initial condition $\tilde{\mathbf{y}}^{(0)}_j=0$, the next iteration gives $\tilde{\mathbf{y}}^{(0)}_j=\mathbf{b}_j$, which is the initial condition from our construction. Therefore, since we are approximating $\mathbf{y}_j=\cC(h\boldsymbol{\Delta})\tilde{\mathbf{y}}_{j-1}$ by $\tilde{\mathbf{y}}_{j}=\tilde{\mathbf{y}}_{j}^{(K)}$, we have
\begin{equation}
\cC(h\boldsymbol{\Delta})\tilde{\mathbf{y}}_{j-1} - \tilde{\mathbf{y}}_{j}=\mathbf{J}^{K+1} \cC(h\boldsymbol{\Delta})\tilde{\mathbf{y}}_{j-1}
\label{eq:11}
\end{equation}

Define the approximation error in $\cC(h\boldsymbol{\Delta})^j\mathbf{f}$ by
\[e_j= \frac{\norm{\cC^j(h\boldsymbol{\Delta})\mathbf{f}-\tilde{\mathbf{y}}_{j}}_2}{\norm{\cC^j(h\boldsymbol{\Delta})\mathbf{f}}_2}.\]
By the triangle inequality, by the fact that $\cC^j(h\boldsymbol{\Delta})$ is unitary, and by (\ref{eq:11})
\begin{equation}
\begin{split}
e_j\leq &\frac{\norm{\cC^j(h\boldsymbol{\Delta})\mathbf{f}-\cC(h\boldsymbol{\Delta})\tilde{\mathbf{y}}_{j-1}}_2}{\norm{\cC^j(h\boldsymbol{\Delta})\mathbf{f}}_2}
+ \frac{\norm{\cC(h\boldsymbol{\Delta})\tilde{\mathbf{y}}_{j-1}-\tilde{\mathbf{y}}_{j}}_2}{\norm{\cC^j(h\boldsymbol{\Delta})\mathbf{f}}_2}\\
=   & \frac{\norm{\cC^{j-1}(h\boldsymbol{\Delta})\mathbf{f}-\tilde{\mathbf{y}}_{j-1}}_2}{\norm{\cC^{j-1}(h\boldsymbol{\Delta})\mathbf{f}}_2}
+ \frac{\norm{\mathbf{J}^{K+1} \cC(h\boldsymbol{\Delta})\tilde{\mathbf{y}}_{j-1}}_2}{\norm{\mathbf{f}}_2}  \\
\leq & e_{j-1} + \norm{\mathbf{J}^{K+1}}_2\frac{\norm{\cC(h\boldsymbol{\Delta})\tilde{\mathbf{y}}_{j-1}}_2}{\norm{\mathbf{f}}_2}\\ 
=  & e_{j-1} + \norm{\mathbf{J}^{K+1}}_2 \frac{\norm{\tilde{\mathbf{y}}_{j-1}}_2}{\norm{\mathbf{f}}_2}  \\
\leq & e_{j-1} + \norm{\mathbf{J}^{K+1}}_2 (1+e_{j-1})
\end{split}
\label{eq:22}
\end{equation}
where the last inequality is due to
\[\begin{split}
\norm{\tilde{\mathbf{y}}_{j-1}}_2 \leq &\norm{\cC^{j-1}(h\boldsymbol{\Delta})\mathbf{f}}_2 + \norm{\cC^{j-1}(h\boldsymbol{\Delta})\mathbf{f}-\tilde{\mathbf{y}}_{j-1}}_2 \\
= & \norm{\mathbf{f}}_2 + \norm{\mathbf{f}}_2 e_{j-1}.
\end{split}\]
Now, using standard norm bounds, in the general case we have $\norm{\mathbf{J}^{K+1}}_2\leq \sqrt{n}\norm{\mathbf{J}^{K+1}}_{\infty}$. Thus, by $\kappa=\norm{\mathbf{J}}_{\infty}$ we have
\[\begin{split}
e_j\leq &  e_{j-1} + \sqrt{n}\norm{\mathbf{J}}^{K+1}_{\infty} (1+e_{j-1}) \\
= & (1+ \sqrt{n}\kappa^{K+1})e_{j-1} + \sqrt{n}\kappa^{K+1}.\end{split}\]
The solution of this recurrent sequence is 
\[e_j\leq(1+\sqrt{n}\kappa^{K+1})^j-1=j\sqrt{n}\kappa^{K+1} + \mathcal{O}(\kappa^{2K+2}). \]
If we use the version of the algorithm, in which each $\tilde{\mathbf{y}}_{j}$ is normalized, we get by (\ref{eq:22})
$e_j\leq e_{j-1}+\sqrt{n}\kappa^{K+1}$. The solution of this recurrent sequence is 
\[e_j\leq j\sqrt{n}\kappa^{K+1}.\] 
We denote in this case $M_j=j\sqrt{n}$

In case the graph is regular, we have $\mathbf{D}=d\mathbf{I}$. In the non-normalized Laplacian case, 
\begin{equation}
\mathbf{J}=-(hd\mathbf{I}+i\mathbf{I})^{-1}h(\boldsymbol{\Delta}-d\mathbf{I}) = \frac{h}{hd+i}(d\mathbf{I}-\boldsymbol{\Delta}) = \frac{h}{hd+i}\mathbf{W}.
\label{eq:33}
\end{equation}
The spectral radius of $\boldsymbol{\Delta}$ is bounded by $2d$.
This can be shown as follows. a value $\lambda$ is not an eigenvalue of $\boldsymbol{\Delta}$ (namely it is a regular value) if and only if $(\boldsymbol{\Delta}-\lambda\mathbf{I})$ is invertible. Moreover, the matrix $(\boldsymbol{\Delta}-\lambda\mathbf{I})$ is strictly dominant diagonal for any $\abs{\lambda} >2d$. By Levy–Desplanques theorem {(\cite{Horn_matrix} Theorem 6.1.10),} any strictly dominant diagonal matrix is invertible, which means that all of the eigenvalues of $\boldsymbol{\Delta}$ are less than $2d$ in their absolute value.
As a result, the spectral radius of $(d\mathbf{I}-\boldsymbol{\Delta})$ is realized on the smallest eigenvalue of $\boldsymbol{\Delta}$, namely it is $\abs{d-0}=d$. This means that the specral radius of $\mathbf{J}$ is
 $\frac{hd}{\sqrt{h^2d^2+1}}$. As a result $\norm{\mathbf{J}}_2=\frac{hd}{\sqrt{h^2d^2+1}}=\kappa$. 
We can now continue from (\ref{eq:22}) to get
\[e_j\leq e_{j-1} + \norm{\mathbf{J}}_2^{K+1} (1+e_{j-1}) = e_{j-1}+\kappa^{K+1}(1+e_{j-1}).\]
As before, we get $e_j\leq j\kappa^{K+1} + \mathcal{O}(\kappa^{2K+2})$, and $e_j\leq j\kappa^{K+1}$ if each $\tilde{\mathbf{y}}_{j}$ is normalized.
We denote in this case $M_j=j$.

In the case of the normalized Laplacian of a regular graph, the spectral radius of $\boldsymbol{\Delta}_n$ is bounded by $2$, and the diagonal entries  are all 1.   
Equation (\ref{eq:33}) in this case reads \newline
 $\mathbf{J} = \frac{h}{h+i}(\mathbf{I}-\boldsymbol{\Delta}_n)$, and $\mathbf{J}$ has spectral radius $\frac{h}{\sqrt{h^2+1}}$. Thus $\norm{\mathbf{J}}_2=\frac{h}{\sqrt{h^2+1}}=\kappa$ and we continue as before to get \newline
 $e_j\leq j\kappa^{K+1}$ and $M_j=j$. 

In all cases, we have by the triangle inequality
\[\begin{split}
\frac{\norm{\mathbf{G}\mathbf{f} - \widetilde{\mathbf{G}\mathbf{f}}}_2}{\norm{\mathbf{f}}_2} 
\leq  &2\sum_{j=1}^r\abs{c_j}\frac{\norm{\cC^j(h\boldsymbol{\Delta})\mathbf{f}-\tilde{\mathbf{y}}_{j}}_2}{\norm{\cC^j(h\boldsymbol{\Delta})\mathbf{f}}_2}\\
 =  &2\sum_{j=1}^r\abs{c_j}e_j \leq 2\sum_{j=1}^rM_j\abs{c_j}\ \kappa^{K+1}.
 \end{split}\]

\section{Proof of Theorem 4}
In this proof we approximate $\mathbf{G}\boldsymbol{\delta}_m$ by $\widetilde{\mathbf{G}\boldsymbol{\delta}}_m$. Note that the signal $\boldsymbol{\delta}_m$ is supported on one vertex, and in the calculation of $\widetilde{\mathbf{G}\boldsymbol{\delta}}_m$, each Jacobi iteration increases the support of the signal by 1-hop. Therefore, the support of $\widetilde{\mathbf{G}\boldsymbol{\delta}}_m$ is the $r(K+1)$-hop neighborhood ${\cal N}_{r(K+1),m}$ of $m$. Denoting $l=r(K+1)$, and using Proposition 1, we get
\begin{multline}
\norm{\mathbf{G}\boldsymbol{\delta}_m - \mathbf{G}\boldsymbol{\delta}_m|_{{\cal N}_{l,m}}}_2\\
 \leq  \norm{\mathbf{G}\boldsymbol{\delta}_m - \widetilde{\mathbf{G}\boldsymbol{\delta}}_m}_2
+\norm{\widetilde{\mathbf{G}\boldsymbol{\delta}}_m - \mathbf{G}\boldsymbol{\delta}_m|_{{\cal N}_{l,m}}}_2\\
 \leq \norm{\mathbf{G}\boldsymbol{\delta}_m - \widetilde{\mathbf{G}\boldsymbol{\delta}}_m}_2
+\norm{\widetilde{\mathbf{G}\boldsymbol{\delta}}_m - \mathbf{G}\boldsymbol{\delta}_m}_2 \\
 = 2\norm{\mathbf{G}\boldsymbol{\delta}_m - \widetilde{\mathbf{G}\boldsymbol{\delta}}_m}_2 \leq 4 M \kappa^{K+1} \norm{\boldsymbol{\delta}_m}_2\\
  = 4 M (\kappa^{1/r})^{l}.
  \label{eq:22}
\end{multline}

\section{Computational Complexity}

In Figure \ref{fig:test-training-times-community-dataset} we compare the computational complexity of CayleyNey and ChebNet on our community detection problem.

%%%%%%%%%%%%%%%%%%COMMON LEGEND%%%%%%%%%%%%%
\newcommand{\realplot}[1]{
\begin{tikzpicture}

       \begin{axis}
        \addlegendimage{empty legend}\addlegendentry{Matrix #1}
        \addplot {0};
        \addplot {1};
        \addplot {2};
        \addplot {3};
        \addplot {4};
        \addplot {5};
    \end{axis}
\end{tikzpicture}}
    % argument #1: any options
    \newenvironment{customlegend}[1][]{%
        \begingroup
        % inits/clears the lists (which might be populated from previous
        % axes):
        \csname pgfplots@init@cleared@structures\endcsname
        \pgfplotsset{#1}%
    }{%
        % draws the legend:
        \csname pgfplots@createlegend\endcsname
        \endgroup
    }%

    % makes \addlegendimage available (typically only available within an
    % axis environment):
    \def\addlegendimage{\csname pgfplots@addlegendimage\endcsname}

%%%%%%%%%%%%%%%%%%%%%%%%%%%%
\begin{figure}[!ht]
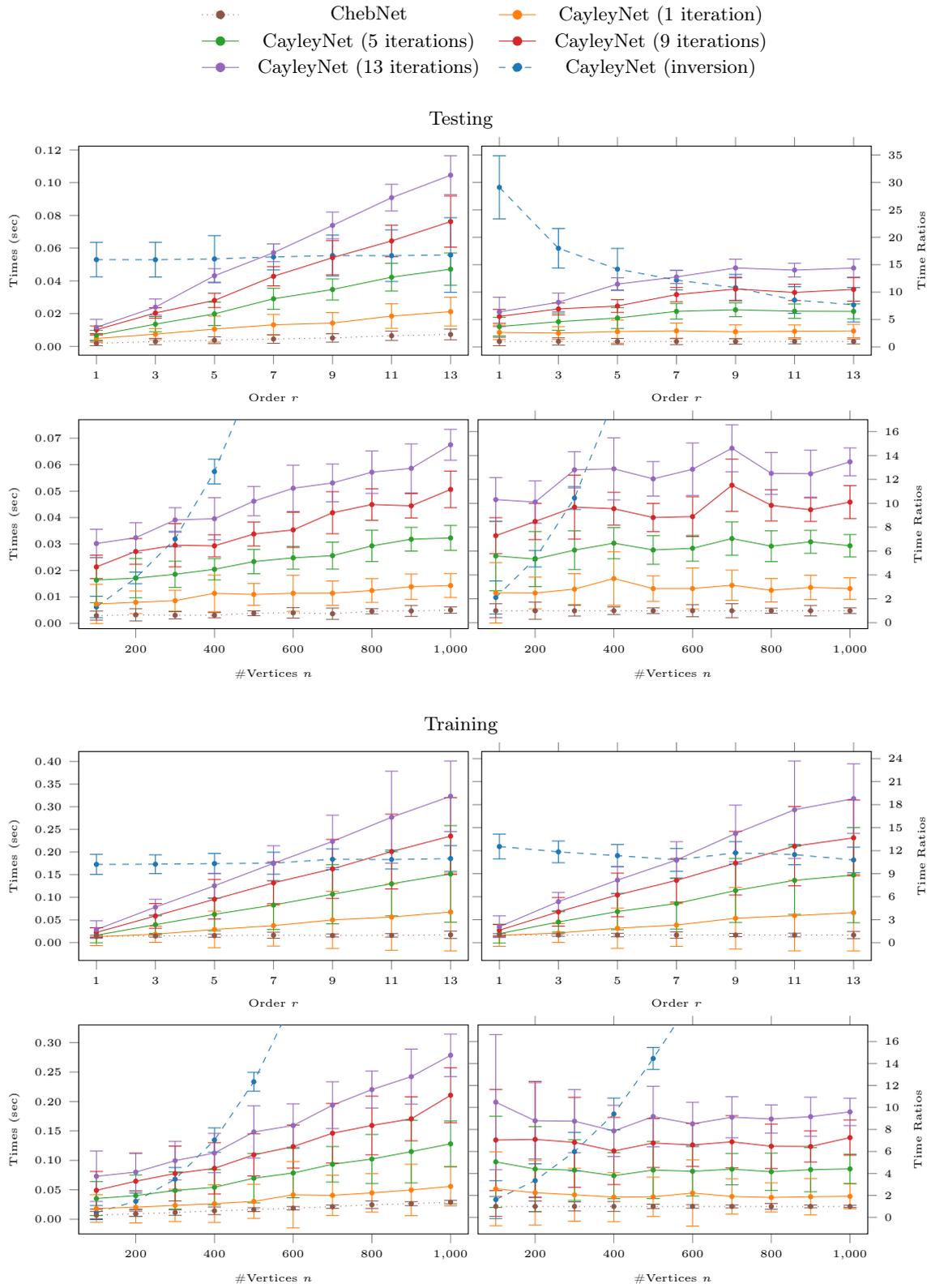

\vspace{-1mm}
\centering 
\setlength\fheight{5cm}
\setlength\fwidth{0.53\linewidth}

    \begin{minipage}{1.0\linewidth}
    \center 
    \hspace{7mm}
   \begin{tikzpicture}
   \definecolor{color1}{rgb}{1,0.498039215686275,0.0549019607843137}
\definecolor{color0}{rgb}{0.12156862745098,0.466666666666667,0.705882352941177}
\definecolor{color3}{rgb}{0.83921568627451,0.152941176470588,0.156862745098039}
\definecolor{color2}{rgb}{0.172549019607843,0.627450980392157,0.172549019607843}
\definecolor{color5}{rgb}{0.549019607843137,0.337254901960784,0.294117647058824}
\definecolor{color4}{rgb}{0.580392156862745,0.403921568627451,0.741176470588235}
        \begin{customlegend}[legend columns=2,legend style={align=left,draw=none,column sep=1.5ex},legend entries={ \small ChebNet, \small CayleyNet (1 iteration),  \small CayleyNet (5 iterations), \small CayleyNet (9 iterations), \small CayleyNet (13 iterations),  \small CayleyNet (inversion)}]%, \small 13 iterations, \small Matrix Inversion }]
        \addlegendimage{mark=*, color5, dotted,line legend}
        \addlegendimage{mark=*, color1, solid}
        \addlegendimage{mark=*, color2, solid}   
        \addlegendimage{mark=*, color3, solid}
        \addlegendimage{mark=*, color4, solid}
        \addlegendimage{mark=*, color0, dashed}
        \end{customlegend}
     \end{tikzpicture} 
          \end{minipage}   
   % \hspace*{2mm}
%%    \begin{minipage}{1.0\linewidth}
%%          \vspace{-2mm} 
%%    \center 
%%      \begin{tikzpicture}
%%   \definecolor{color1}{rgb}{1,0.498039215686275,0.0549019607843137}
%%\definecolor{color0}{rgb}{0.12156862745098,0.466666666666667,0.705882352941177}
%%\definecolor{color3}{rgb}{0.83921568627451,0.152941176470588,0.156862745098039}
%%\definecolor{color2}{rgb}{0.172549019607843,0.627450980392157,0.172549019607843}
%%\definecolor{color5}{rgb}{0.549019607843137,0.337254901960784,0.294117647058824}
%%\definecolor{color4}{rgb}{0.580392156862745,0.403921568627451,0.741176470588235}
%%        \begin{customlegend}[legend columns=2,legend style={align=left,draw=none,column sep=1.5ex, fill=none},legend entries={ CayleyNet (9 iterations), CayleyNet (13 iterations),  CayleyNet (inversion)}]
%%%        \addlegendimage{mark=*, color5, dotted,line legend}
%%%        \addlegendimage{mark=*, color1, solid}
%%%        \addlegendimage{mark=*, color2, solid}   
%%        \addlegendimage{mark=*, color3, solid,line legend}
%%        \addlegendimage{mark=*, color4, solid}
%%        \addlegendimage{mark=*, color0, dashed}
%%        \end{customlegend}
%%     \end{tikzpicture} 
%%
%%     \end{minipage}
%\noindent\fcolorbox{white!92!gray}{white!92!gray}{
\begin{minipage}{1.0\linewidth}
\centering
\vspace*{4mm}
{\small Testing}\\
\vspace*{1mm}
\begin{minipage}{1.0\linewidth}
% This file was created by matplotlib2tikz v0.6.7.
\begin{tikzpicture}[font=\tiny]

\definecolor{color1}{rgb}{1,0.498039215686275,0.0549019607843137}
\definecolor{color0}{rgb}{0.12156862745098,0.466666666666667,0.705882352941177}
\definecolor{color3}{rgb}{0.83921568627451,0.152941176470588,0.156862745098039}
\definecolor{color2}{rgb}{0.172549019607843,0.627450980392157,0.172549019607843}
\definecolor{color5}{rgb}{0.549019607843137,0.337254901960784,0.294117647058824}
\definecolor{color4}{rgb}{0.580392156862745,0.403921568627451,0.741176470588235}

\begin{axis}[
xlabel={\tiny Order $r$},
ylabel={\tiny Times (sec)},
xmin=0.4, xmax=13.6,
ymin=-0.00537421617610107, ymax=0.122265122212567,
yticklabel style={
        /pgf/number format/fixed,
        /pgf/number format/fixed zerofill,
        /pgf/number format/precision=2
},
xtick={1,3,5,7,9,11,13},
ytick={0, 0.02, 0.04, 0.06, 0.08, 0.1, 0.12},
width=\fwidth,
height=\fheight,
tick align=outside,
tick pos=left,
x grid style={lightgray!92.026143790849673!black},
y grid style={lightgray!92.026143790849673!black},
%legend entries={{Matrix inversion},{No. Jacobi iter 1},{No. Jacobi iter 5},{No. Jacobi iter 9},{No. Jacobi iter 13},{ChebNet}},
%legend cell align={left},
%legend style={at={(0.03,0.97)}, anchor=north west, draw=white!80.0!black}
]
\path [draw=color0, thin] (axis cs:1,0.0424847443830264)
--(axis cs:1,0.0634809970923967);

\path [draw=color0, thin] (axis cs:3,0.0423322032412943)
--(axis cs:3,0.0635047922014141);

\path [draw=color0, thin] (axis cs:5,0.0391510460876592)
--(axis cs:5,0.0676549142496618);

\path [draw=color0, thin] (axis cs:7,0.0466586989256406)
--(axis cs:7,0.0625701028016543);

\path [draw=color0, thin] (axis cs:9,0.0429939448208937)
--(axis cs:9,0.0679212233055304);

\path [draw=color0, thin] (axis cs:11,0.0396157061904174)
--(axis cs:11,0.0711649461736776);

\path [draw=color0, thin] (axis cs:13,0.0330291125008533)
--(axis cs:13,0.0786092268597017);

\path [draw=color1, thin] (axis cs:1,0.00243402464270362)
--(axis cs:1,0.00720622397383284);

\path [draw=color1, thin] (axis cs:3,0.00405003860241846)
--(axis cs:3,0.0108340359265936);

\path [draw=color1, thin] (axis cs:5,0.00260430514100481)
--(axis cs:5,0.0184681938163113);

\path [draw=color1, thin] (axis cs:7,0.00680923436354451)
--(axis cs:7,0.0194749039980684);

\path [draw=color1, thin] (axis cs:9,0.00783598808696207)
--(axis cs:9,0.0206027389739917);

\path [draw=color1, thin] (axis cs:11,0.0109803394738373)
--(axis cs:11,0.0259766860541168);

\path [draw=color1, thin] (axis cs:13,0.0124095008898262)
--(axis cs:13,0.0299431119234876);

\path [draw=color2, thin] (axis cs:1,0.0036792372676835)
--(axis cs:1,0.00982383737831259);

\path [draw=color2, thin] (axis cs:3,0.0089283645532845)
--(axis cs:3,0.0181728978571973);

\path [draw=color2, thin] (axis cs:5,0.0126718886969604)
--(axis cs:5,0.0269717486423137);

\path [draw=color2, thin] (axis cs:7,0.0226466393505617)
--(axis cs:7,0.0354855481748696);

\path [draw=color2, thin] (axis cs:9,0.02827395785713)
--(axis cs:9,0.0411512117697417);

\path [draw=color2, thin] (axis cs:11,0.0337450925975795)
--(axis cs:11,0.0507121141284947);

\path [draw=color2, thin] (axis cs:13,0.037281213309535)
--(axis cs:13,0.0569765864748075);

\path [draw=color3, thin] (axis cs:1,0.00781430713371446)
--(axis cs:1,0.0124316740827162);

\path [draw=color3, thin] (axis cs:3,0.0172055822326119)
--(axis cs:3,0.0235592582113171);

\path [draw=color3, thin] (axis cs:5,0.0237302922946833)
--(axis cs:5,0.0324066496151067);

\path [draw=color3, thin] (axis cs:7,0.0369042122841494)
--(axis cs:7,0.0486389274914446);

\path [draw=color3, thin] (axis cs:9,0.0436596576575494)
--(axis cs:9,0.0646449859734321);

\path [draw=color3, thin] (axis cs:11,0.0547543488219527)
--(axis cs:11,0.0740228372538936);

\path [draw=color3, thin] (axis cs:13,0.0605630757561103)
--(axis cs:13,0.0918858963101332);

\path [draw=color4, thin] (axis cs:1,0.00674728340098467)
--(axis cs:1,0.0164749564112305);

\path [draw=color4, thin] (axis cs:3,0.0189824168672536)
--(axis cs:3,0.0289201036303864);

\path [draw=color4, thin] (axis cs:5,0.0387642338011772)
--(axis cs:5,0.0474762008454293);

\path [draw=color4, thin] (axis cs:7,0.0517954888504718)
--(axis cs:7,0.0625953612166669);

\path [draw=color4, thin] (axis cs:9,0.065715300781323)
--(axis cs:9,0.0820474269561037);

\path [draw=color4, thin] (axis cs:11,0.0826475677361731)
--(axis cs:11,0.0989690882493412);

\path [draw=color4, thin] (axis cs:13,0.0926611674422323)
--(axis cs:13,0.116463334103992);

\path [draw=color5, thin] (axis cs:1,0.000427571932474772)
--(axis cs:1,0.00321454207102458);

\path [draw=color5, thin] (axis cs:3,0.000997281074523926)
--(axis cs:3,0.00488817691802979);

\path [draw=color5, thin] (axis cs:5,0.00168119271596273)
--(axis cs:5,0.00585173765818278);

\path [draw=color5, thin] (axis cs:7,0.00186629295349121)
--(axis cs:7,0.0070990244547526);

\path [draw=color5, thin] (axis cs:9,0.00257449150085449)
--(axis cs:9,0.00768725077311198);

\path [draw=color5, thin] (axis cs:11,0.00361653963724772)
--(axis cs:11,0.00935239791870117);

\path [draw=color5, thin] (axis cs:13,0.00399941603342692)
--(axis cs:13,0.010542639096578);

\addplot [thin, color0, mark=-, mark size=3, mark options={solid}, only marks]
table {%
1 0.0424847443830264
3 0.0423322032412943
5 0.0391510460876592
7 0.0466586989256406
9 0.0429939448208937
11 0.0396157061904174
13 0.0330291125008533
};
\addplot [thin, color0, mark=-, mark size=3, mark options={solid}, only marks]
table {%
1 0.0634809970923967
3 0.0635047922014141
5 0.0676549142496618
7 0.0625701028016543
9 0.0679212233055304
11 0.0711649461736776
13 0.0786092268597017
};
\addplot [thin, color1, mark=-, mark size=3, mark options={solid}, only marks]
table {%
1 0.00243402464270362
3 0.00405003860241846
5 0.00260430514100481
7 0.00680923436354451
9 0.00783598808696207
11 0.0109803394738373
13 0.0124095008898262
};
\addplot [thin, color1, mark=-, mark size=3, mark options={solid}, only marks]
table {%
1 0.00720622397383284
3 0.0108340359265936
5 0.0184681938163113
7 0.0194749039980684
9 0.0206027389739917
11 0.0259766860541168
13 0.0299431119234876
};
\addplot [thin, color2, mark=-, mark size=3, mark options={solid}, only marks]
table {%
1 0.0036792372676835
3 0.0089283645532845
5 0.0126718886969604
7 0.0226466393505617
9 0.02827395785713
11 0.0337450925975795
13 0.037281213309535
};
\addplot [thin, color2, mark=-, mark size=3, mark options={solid}, only marks]
table {%
1 0.00982383737831259
3 0.0181728978571973
5 0.0269717486423137
7 0.0354855481748696
9 0.0411512117697417
11 0.0507121141284947
13 0.0569765864748075
};
\addplot [thin, color3, mark=-, mark size=3, mark options={solid}, only marks]
table {%
1 0.00781430713371446
3 0.0172055822326119
5 0.0237302922946833
7 0.0369042122841494
9 0.0436596576575494
11 0.0547543488219527
13 0.0605630757561103
};
\addplot [thin, color3, mark=-, mark size=3, mark options={solid}, only marks]
table {%
1 0.0124316740827162
3 0.0235592582113171
5 0.0324066496151067
7 0.0486389274914446
9 0.0646449859734321
11 0.0740228372538936
13 0.0918858963101332
};
\addplot [thin, color4, mark=-, mark size=3, mark options={solid}, only marks]
table {%
1 0.00674728340098467
3 0.0189824168672536
5 0.0387642338011772
7 0.0517954888504718
9 0.065715300781323
11 0.0826475677361731
13 0.0926611674422323
};
\addplot [thin, color4, mark=-, mark size=3, mark options={solid}, only marks]
table {%
1 0.0164749564112305
3 0.0289201036303864
5 0.0474762008454293
7 0.0625953612166669
9 0.0820474269561037
11 0.0989690882493412
13 0.116463334103992
};
\addplot [thin, color5, mark=-, mark size=3, mark options={solid}, only marks]
table {%
1 0.000427571932474772
3 0.000997281074523926
5 0.00168119271596273
7 0.00186629295349121
9 0.00257449150085449
11 0.00361653963724772
13 0.00399941603342692
};
\addplot [thin, color5, mark=-, mark size=3, mark options={solid}, only marks]
table {%
1 0.00321454207102458
3 0.00488817691802979
5 0.00585173765818278
7 0.0070990244547526
9 0.00768725077311198
11 0.00935239791870117
13 0.010542639096578
};
\addplot [thin, color0, dashed, mark=*, mark size=1, mark options={solid}]
table {%
1 0.0529828707377116
3 0.0529184977213542
5 0.0534029801686605
7 0.0546144008636475
9 0.0554575840632121
11 0.0553903261820475
13 0.0558191696802775
};
\addplot [thin, color1, mark=*, mark size=1, mark options={solid}]
table {%
1 0.00482012430826823
3 0.00744203726450602
5 0.010536249478658
7 0.0131420691808065
9 0.0142193635304769
11 0.0184785127639771
13 0.0211763064066569
};
\addplot [thin, color2, mark=*, mark size=1, mark options={solid}]
table {%
1 0.00675153732299805
3 0.0135506312052409
5 0.019821818669637
7 0.0290660937627157
9 0.0347125848134359
11 0.0422286033630371
13 0.0471288998921712
};
\addplot [thin, color3, mark=*, mark size=1, mark options={solid}]
table {%
1 0.0101229906082153
3 0.0203824202219645
5 0.028068470954895
7 0.042771569887797
9 0.0541523218154907
11 0.0643885930379232
13 0.0762244860331217
};
\addplot [thin, color4, mark=*, mark size=1, mark options={solid}]
table {%
1 0.0116111199061076
3 0.02395126024882
5 0.0431202173233032
7 0.0571954250335693
9 0.0738813638687134
11 0.0908083279927572
13 0.104562250773112
};
\addplot [thin, color5, dotted, mark=*, mark size=1, mark options={solid}]
table {%
1 0.00182105700174967
3 0.00294272899627686
5 0.00376646518707275
7 0.00448265870412191
9 0.00513087113698324
11 0.00648446877797445
13 0.00727102756500244
};
\end{axis}

\end{tikzpicture}%\hfill
\hspace*{-1mm} 
% This file was created by matplotlib2tikz v0.6.7.
\begin{tikzpicture}[font=\tiny]

\definecolor{color1}{rgb}{1,0.498039215686275,0.0549019607843137}
\definecolor{color0}{rgb}{0.12156862745098,0.466666666666667,0.705882352941177}
\definecolor{color3}{rgb}{0.83921568627451,0.152941176470588,0.156862745098039}
\definecolor{color2}{rgb}{0.172549019607843,0.627450980392157,0.172549019607843}
\definecolor{color5}{rgb}{0.549019607843137,0.337254901960784,0.294117647058824}
\definecolor{color4}{rgb}{0.580392156862745,0.403921568627451,0.741176470588235}

\begin{axis}[
xlabel={\tiny Order $r$},
ylabel={\tiny Time Ratios},
xmin=0.4, xmax=13.6,
ymin=-1.49643823499377, ymax=36.5906549253379,
xtick={1,3,5,7,9,11,13},
ytick={0, 5, 10, 15, 20, 25, 30, 35},
width=\fwidth,
height=\fheight,
tick align=outside,
%tick pos=left,
yticklabel pos=right,
x grid style={lightgray!92.026143790849673!black},
y grid style={lightgray!92.026143790849673!black},
%legend style={draw=white!80.0!black},
%legend entries={{Matrix inversion},{No. Jacobi iter 1},{No. Jacobi iter 5},{No. Jacobi iter 9},{No. Jacobi iter 13},{ChebNet}},
%legend cell align={left}
]
\path [draw=color0, thin] (axis cs:1,23.3297169403303)
--(axis cs:1,34.8594234180501);

\path [draw=color0, thin] (axis cs:3,14.385355666408)
--(axis cs:3,21.5802380313513);

\path [draw=color0, thin] (axis cs:5,10.3946390430033)
--(axis cs:5,17.9624424730824);

\path [draw=color0, thin] (axis cs:7,10.4087109917016)
--(axis cs:7,13.9582571263165);

\path [draw=color0, thin] (axis cs:9,8.37946299430404)
--(axis cs:9,13.2377566093904);

\path [draw=color0, thin] (axis cs:11,6.10932175739339)
--(axis cs:11,10.9746763552014);

\path [draw=color0, thin] (axis cs:13,4.54256461078927)
--(axis cs:13,10.8112953990259);

\path [draw=color1, thin] (axis cs:1,1.33659991991739)
--(axis cs:1,3.95716551810794);

\path [draw=color1, thin] (axis cs:3,1.37628663989874)
--(axis cs:3,3.68162883510538);

\path [draw=color1, thin] (axis cs:5,0.691445430039627)
--(axis cs:5,4.90332258471357);

\path [draw=color1, thin] (axis cs:7,1.5190169078193)
--(axis cs:7,4.34449849598427);

\path [draw=color1, thin] (axis cs:9,1.52722371654988)
--(axis cs:9,4.01544658284);

\path [draw=color1, thin] (axis cs:11,1.69332906823976)
--(axis cs:11,4.00598521537351);

\path [draw=color1, thin] (axis cs:13,1.70670524611359)
--(axis cs:13,4.1181403392847);

\path [draw=color2, thin] (axis cs:1,2.02038555857861)
--(axis cs:1,5.39457983406002);

\path [draw=color2, thin] (axis cs:3,3.03404240233494)
--(axis cs:3,6.17552546639179);

\path [draw=color2, thin] (axis cs:5,3.36439820032131)
--(axis cs:5,7.16102427679034);

\path [draw=color2, thin] (axis cs:7,5.05205523002178)
--(axis cs:7,7.91618334499564);

\path [draw=color2, thin] (axis cs:9,5.51055699944046)
--(axis cs:9,8.02031675929735);

\path [draw=color2, thin] (axis cs:11,5.20398721205972)
--(axis cs:11,7.82055028173574);

\path [draw=color2, thin] (axis cs:13,5.12736514560614)
--(axis cs:13,7.83611201655351);

\path [draw=color3, thin] (axis cs:1,4.29108321497156)
--(axis cs:1,6.82662545476162);

\path [draw=color3, thin] (axis cs:3,5.84681166848202)
--(axis cs:3,8.00592179610298);

\path [draw=color3, thin] (axis cs:5,6.30041461052934)
--(axis cs:5,8.60399552512331);

\path [draw=color3, thin] (axis cs:7,8.23266162338328)
--(axis cs:7,10.8504641334215);

\path [draw=color3, thin] (axis cs:9,8.50920954589003)
--(axis cs:9,12.5992222855632);

\path [draw=color3, thin] (axis cs:11,8.44392203844589)
--(axis cs:11,11.4154049912807);

\path [draw=color3, thin] (axis cs:13,8.32936957186325)
--(axis cs:13,12.6372641952847);

\path [draw=color4, thin] (axis cs:1,3.7051467331895)
--(axis cs:1,9.04691967104892);

\path [draw=color4, thin] (axis cs:3,6.45061672048977)
--(axis cs:3,9.82764762469672);

\path [draw=color4, thin] (axis cs:5,10.2919400222319)
--(axis cs:5,12.6049753515251);

\path [draw=color4, thin] (axis cs:7,11.5546358242363)
--(axis cs:7,13.9638918214115);

\path [draw=color4, thin] (axis cs:9,12.807825226334)
--(axis cs:9,15.9909350216783);

\path [draw=color4, thin] (axis cs:11,12.7454646735133)
--(axis cs:11,15.2624820379282);

\path [draw=color4, thin] (axis cs:13,12.7438888951869)
--(axis cs:13,16.0174518749679);

\path [draw=color5, thin] (axis cs:1,0.234793272294036)
--(axis cs:1,1.76520672770596);

\path [draw=color5, thin] (axis cs:3,0.338896675767863)
--(axis cs:3,1.66110332423214);

\path [draw=color5, thin] (axis cs:5,0.446358225142479)
--(axis cs:5,1.55364177485752);

\path [draw=color5, thin] (axis cs:7,0.416336169375356)
--(axis cs:7,1.58366383062464);

\path [draw=color5, thin] (axis cs:9,0.501764989242782)
--(axis cs:9,1.49823501075722);

\path [draw=color5, thin] (axis cs:11,0.557723348060814)
--(axis cs:11,1.44227665193919);

\path [draw=color5, thin] (axis cs:13,0.550048256270856)
--(axis cs:13,1.44995174372914);

\addplot [thin, color0, mark=-, mark size=3, mark options={solid}, only marks]
table {%
1 23.3297169403303
3 14.385355666408
5 10.3946390430033
7 10.4087109917016
9 8.37946299430404
11 6.10932175739339
13 4.54256461078927
};
\addplot [thin, color0, mark=-, mark size=3, mark options={solid}, only marks]
table {%
1 34.8594234180501
3 21.5802380313513
5 17.9624424730824
7 13.9582571263165
9 13.2377566093904
11 10.9746763552014
13 10.8112953990259
};
\addplot [thin, color1, mark=-, mark size=3, mark options={solid}, only marks]
table {%
1 1.33659991991739
3 1.37628663989874
5 0.691445430039627
7 1.5190169078193
9 1.52722371654988
11 1.69332906823976
13 1.70670524611359
};
\addplot [thin, color1, mark=-, mark size=3, mark options={solid}, only marks]
table {%
1 3.95716551810794
3 3.68162883510538
5 4.90332258471357
7 4.34449849598427
9 4.01544658284
11 4.00598521537351
13 4.1181403392847
};
\addplot [thin, color2, mark=-, mark size=3, mark options={solid}, only marks]
table {%
1 2.02038555857861
3 3.03404240233494
5 3.36439820032131
7 5.05205523002178
9 5.51055699944046
11 5.20398721205972
13 5.12736514560614
};
\addplot [thin, color2, mark=-, mark size=3, mark options={solid}, only marks]
table {%
1 5.39457983406002
3 6.17552546639179
5 7.16102427679034
7 7.91618334499564
9 8.02031675929735
11 7.82055028173574
13 7.83611201655351
};
\addplot [thin, color3, mark=-, mark size=3, mark options={solid}, only marks]
table {%
1 4.29108321497156
3 5.84681166848202
5 6.30041461052934
7 8.23266162338328
9 8.50920954589003
11 8.44392203844589
13 8.32936957186325
};
\addplot [thin, color3, mark=-, mark size=3, mark options={solid}, only marks]
table {%
1 6.82662545476162
3 8.00592179610298
5 8.60399552512331
7 10.8504641334215
9 12.5992222855632
11 11.4154049912807
13 12.6372641952847
};
\addplot [thin, color4, mark=-, mark size=3, mark options={solid}, only marks]
table {%
1 3.7051467331895
3 6.45061672048977
5 10.2919400222319
7 11.5546358242363
9 12.807825226334
11 12.7454646735133
13 12.7438888951869
};
\addplot [thin, color4, mark=-, mark size=3, mark options={solid}, only marks]
table {%
1 9.04691967104892
3 9.82764762469672
5 12.6049753515251
7 13.9638918214115
9 15.9909350216783
11 15.2624820379282
13 16.0174518749679
};
\addplot [thin, color5, mark=-, mark size=3, mark options={solid}, only marks]
table {%
1 0.234793272294036
3 0.338896675767863
5 0.446358225142479
7 0.416336169375356
9 0.501764989242782
11 0.557723348060814
13 0.550048256270856
};
\addplot [thin, color5, mark=-, mark size=3, mark options={solid}, only marks]
table {%
1 1.76520672770596
3 1.66110332423214
5 1.55364177485752
7 1.58366383062464
9 1.49823501075722
11 1.44227665193919
13 1.44995174372914
};
\addplot [thin, color0, dashed, mark=*, mark size=1, mark options={solid}]
table {%
1 29.0945701791902
3 17.9827968488796
5 14.1785407580428
7 12.1834840590091
9 10.8086098018472
11 8.54199905629738
13 7.6769300049076
};
\addplot [thin, color1, mark=*, mark size=1, mark options={solid}]
table {%
1 2.64688271901266
3 2.52895773750206
5 2.7973840073766
7 2.93175770190178
9 2.77133514969494
11 2.84965714180664
13 2.91242279269915
};
\addplot [thin, color2, mark=*, mark size=1, mark options={solid}]
table {%
1 3.70748269631931
3 4.60478393436336
5 5.26271123855582
7 6.48411928750871
9 6.76543687936891
11 6.51226874689773
13 6.48173858107983
};
\addplot [thin, color3, mark=*, mark size=1, mark options={solid}]
table {%
1 5.55885433486659
3 6.9263667322925
5 7.45220506782633
7 9.54156287840241
9 10.5542159157266
11 9.92966351486332
13 10.483316883574
};
\addplot [thin, color4, mark=*, mark size=1, mark options={solid}]
table {%
1 6.37603320211921
3 8.13913217259325
5 11.4484576868785
7 12.7592638228239
9 14.3993801240062
11 14.0039733557207
13 14.3806703850774
};
\addplot [thin, color5, dotted, mark=*, mark size=1, mark options={solid}]
table {%
1 1
3 1
5 1
7 1
9 1
11 1
13 1
};
\end{axis}

\end{tikzpicture}%\hfill
\end{minipage}\\
\begin{minipage}{1.0\linewidth}
\input{figures/community_detection/new_times/CayleyNet_test_time_inv_jacobi_order_batch_analysis_multiple_size_max_1000vert_sparse_with_std.tex}%\hfill 
\hspace*{-2.6mm} 
\input{figures/community_detection/new_times/CayleyNet_test_time_inv_jacobi_order_batch_analysis_multiple_size_max_1000vert_sparse_ratio_with_std.tex}%\hfill
\end{minipage}
\vspace{-2mm}
\end{minipage}
%}
\vspace{1mm}
\begin{minipage}{1.0\linewidth}

\vspace*{5mm}
\centering
{\small Training}\\
\vspace*{1mm}
\begin{minipage}{1.0\linewidth}
% This file was created by matplotlib2tikz v0.6.7.
\begin{tikzpicture}[font=\tiny]

\definecolor{color1}{rgb}{1,0.498039215686275,0.0549019607843137}
\definecolor{color0}{rgb}{0.12156862745098,0.466666666666667,0.705882352941177}
\definecolor{color3}{rgb}{0.83921568627451,0.152941176470588,0.156862745098039}
\definecolor{color2}{rgb}{0.172549019607843,0.627450980392157,0.172549019607843}
\definecolor{color5}{rgb}{0.549019607843137,0.337254901960784,0.294117647058824}
\definecolor{color4}{rgb}{0.580392156862745,0.403921568627451,0.741176470588235}

\begin{axis}[
xlabel={\tiny Order $r$},
ylabel={\tiny Times (sec)},
xmin=0.4, xmax=13.6,
ymin=-0.0393948091417996, ymax=0.422194609996167,
width=\fwidth,
height=\fheight,
tick align=outside,
tick pos=left,
x grid style={lightgray!92.026143790849673!black},
y grid style={lightgray!92.026143790849673!black},
xtick={1, 3, 5, 7, 9, 11, 13},
ytick={0, 0.05, 0.1, 0.15, 0.2, 0.25, 0.3, 0.35, 0.4},
yticklabel style={
        /pgf/number format/fixed,
        /pgf/number format/fixed zerofill,
        /pgf/number format/precision=2
},
scaled y ticks=false,
%legend entries={{Matrix inversion},{No. Jacobi iter 1},{No. Jacobi iter 5},{No. Jacobi iter 9},{No. Jacobi iter 13},{ChebNet}},
%legend cell align={left},
%legend style={at={(0.03,0.97)}, anchor=north west, draw=white!80.0!black}
]
\path [draw=color0, thin] (axis cs:1,0.150559722254499)
--(axis cs:1,0.194904125859515);

\path [draw=color0, thin] (axis cs:3,0.152465795732941)
--(axis cs:3,0.193965014559621);

\path [draw=color0, thin] (axis cs:5,0.152164584331349)
--(axis cs:5,0.196924783853059);

\path [draw=color0, thin] (axis cs:7,0.151162610362756)
--(axis cs:7,0.19964465905151);

\path [draw=color0, thin] (axis cs:9,0.16118989373168)
--(axis cs:9,0.206850365947479);

\path [draw=color0, thin] (axis cs:11,0.162455717318938)
--(axis cs:11,0.20450573007893);

\path [draw=color0, thin] (axis cs:13,0.157088539658155)
--(axis cs:13,0.214161914926603);

\path [draw=color1, thin] (axis cs:1,-0.00653945585498581)
--(axis cs:1,0.0327798046709038);

\path [draw=color1, thin] (axis cs:3,0.000617433942111074)
--(axis cs:3,0.0363173391851269);

\path [draw=color1, thin] (axis cs:5,-0.0113366654989906)
--(axis cs:5,0.0694386533377357);

\path [draw=color1, thin] (axis cs:7,-0.00771627638391879)
--(axis cs:7,0.0826111496851069);

\path [draw=color1, thin] (axis cs:9,-0.0129585358408158)
--(axis cs:9,0.113013674110876);

\path [draw=color1, thin] (axis cs:11,-0.0169733946465031)
--(axis cs:11,0.129853863015033);

\path [draw=color1, thin] (axis cs:13,-0.0184134719082557)
--(axis cs:13,0.153250415852551);

\path [draw=color2, thin] (axis cs:1,-0.000491230396074075)
--(axis cs:1,0.0329020109848599);

\path [draw=color2, thin] (axis cs:3,0.0201747856926405)
--(axis cs:3,0.0589027124254422);

\path [draw=color2, thin] (axis cs:5,0.0259686897632076)
--(axis cs:5,0.0988358864747888);

\path [draw=color2, thin] (axis cs:7,0.0288712924645141)
--(axis cs:7,0.137050777085046);

\path [draw=color2, thin] (axis cs:9,0.0413719017821162)
--(axis cs:9,0.172300148058175);

\path [draw=color2, thin] (axis cs:11,0.0589926490170703)
--(axis cs:11,0.200823059779113);

\path [draw=color2, thin] (axis cs:13,0.0449401851557088)
--(axis cs:13,0.258443896366915);

\path [draw=color3, thin] (axis cs:1,0.0121324956505527)
--(axis cs:1,0.0327638685614834);

\path [draw=color3, thin] (axis cs:3,0.0315534927658532)
--(axis cs:3,0.0861412666030433);

\path [draw=color3, thin] (axis cs:5,0.0520021375268551)
--(axis cs:5,0.139341265335694);

\path [draw=color3, thin] (axis cs:7,0.0857965376270978)
--(axis cs:7,0.178175665309711);

\path [draw=color3, thin] (axis cs:9,0.09759931858603)
--(axis cs:9,0.228036353029571);

\path [draw=color3, thin] (axis cs:11,0.118443862715602)
--(axis cs:11,0.283674549938003);

\path [draw=color3, thin] (axis cs:13,0.150282823449004)
--(axis cs:13,0.320204421474905);

\path [draw=color4, thin] (axis cs:1,0.00927246269532082)
--(axis cs:1,0.0481413537567055);

\path [draw=color4, thin] (axis cs:3,0.0608465008777697)
--(axis cs:3,0.0957546896892469);

\path [draw=color4, thin] (axis cs:5,0.0978689513690252)
--(axis cs:5,0.153050708563429);

\path [draw=color4, thin] (axis cs:7,0.13599471006601)
--(axis cs:7,0.213967063488474);

\path [draw=color4, thin] (axis cs:9,0.165490203287662)
--(axis cs:9,0.281556235565284);

\path [draw=color4, thin] (axis cs:11,0.175349624669004)
--(axis cs:11,0.378567465429059);

\path [draw=color4, thin] (axis cs:13,0.245159999179516)
--(axis cs:13,0.401213272762623);

\path [draw=color5, thin] (axis cs:1,0.0113582395630838)
--(axis cs:1,0.0161844945830344);

\path [draw=color5, thin] (axis cs:3,0.0121424114650791)
--(axis cs:3,0.017107193077971);

\path [draw=color5, thin] (axis cs:5,0.0118419977939367)
--(axis cs:5,0.018909707375741);

\path [draw=color5, thin] (axis cs:7,0.00960324179593173)
--(axis cs:7,0.0228853904062167);

\path [draw=color5, thin] (axis cs:9,0.0123537659057849)
--(axis cs:9,0.0190530658355481);

\path [draw=color5, thin] (axis cs:11,0.0121077084437616)
--(axis cs:11,0.0198491867487025);

\path [draw=color5, thin] (axis cs:13,0.0091241010421059)
--(axis cs:13,0.0252914142217694);

\addplot [thin, color0, mark=-, mark size=3, mark options={solid}, only marks]
table {%
1 0.150559722254499
3 0.152465795732941
5 0.152164584331349
7 0.151162610362756
9 0.16118989373168
11 0.162455717318938
13 0.157088539658155
};
\addplot [thin, color0, mark=-, mark size=3, mark options={solid}, only marks]
table {%
1 0.194904125859515
3 0.193965014559621
5 0.196924783853059
7 0.19964465905151
9 0.206850365947479
11 0.20450573007893
13 0.214161914926603
};
\addplot [thin, color1, mark=-, mark size=3, mark options={solid}, only marks]
table {%
1 -0.00653945585498581
3 0.000617433942111074
5 -0.0113366654989906
7 -0.00771627638391879
9 -0.0129585358408158
11 -0.0169733946465031
13 -0.0184134719082557
};
\addplot [thin, color1, mark=-, mark size=3, mark options={solid}, only marks]
table {%
1 0.0327798046709038
3 0.0363173391851269
5 0.0694386533377357
7 0.0826111496851069
9 0.113013674110876
11 0.129853863015033
13 0.153250415852551
};
\addplot [thin, color2, mark=-, mark size=3, mark options={solid}, only marks]
table {%
1 -0.000491230396074075
3 0.0201747856926405
5 0.0259686897632076
7 0.0288712924645141
9 0.0413719017821162
11 0.0589926490170703
13 0.0449401851557088
};
\addplot [thin, color2, mark=-, mark size=3, mark options={solid}, only marks]
table {%
1 0.0329020109848599
3 0.0589027124254422
5 0.0988358864747888
7 0.137050777085046
9 0.172300148058175
11 0.200823059779113
13 0.258443896366915
};
\addplot [thin, color3, mark=-, mark size=3, mark options={solid}, only marks]
table {%
1 0.0121324956505527
3 0.0315534927658532
5 0.0520021375268551
7 0.0857965376270978
9 0.09759931858603
11 0.118443862715602
13 0.150282823449004
};
\addplot [thin, color3, mark=-, mark size=3, mark options={solid}, only marks]
table {%
1 0.0327638685614834
3 0.0861412666030433
5 0.139341265335694
7 0.178175665309711
9 0.228036353029571
11 0.283674549938003
13 0.320204421474905
};
\addplot [thin, color4, mark=-, mark size=3, mark options={solid}, only marks]
table {%
1 0.00927246269532082
3 0.0608465008777697
5 0.0978689513690252
7 0.13599471006601
9 0.165490203287662
11 0.175349624669004
13 0.245159999179516
};
\addplot [thin, color4, mark=-, mark size=3, mark options={solid}, only marks]
table {%
1 0.0481413537567055
3 0.0957546896892469
5 0.153050708563429
7 0.213967063488474
9 0.281556235565284
11 0.378567465429059
13 0.401213272762623
};
\addplot [thin, color5, mark=-, mark size=3, mark options={solid}, only marks]
table {%
1 0.0113582395630838
3 0.0121424114650791
5 0.0118419977939367
7 0.00960324179593173
9 0.0123537659057849
11 0.0121077084437616
13 0.0091241010421059
};
\addplot [thin, color5, mark=-, mark size=3, mark options={solid}, only marks]
table {%
1 0.0161844945830344
3 0.017107193077971
5 0.018909707375741
7 0.0228853904062167
9 0.0190530658355481
11 0.0198491867487025
13 0.0252914142217694
};
\addplot [thin, color0, dashed, mark=*, mark size=1, mark options={solid}]
table {%
1 0.172731924057007
3 0.173215405146281
5 0.174544684092204
7 0.175403634707133
9 0.184020129839579
11 0.183480723698934
13 0.185625227292379
};
\addplot [thin, color1, mark=*, mark size=1, mark options={solid}]
table {%
1 0.013120174407959
3 0.018467386563619
5 0.0290509939193726
7 0.0374474366505941
9 0.0500275691350301
11 0.0564402341842651
13 0.0674184719721476
};
\addplot [thin, color2, mark=*, mark size=1, mark options={solid}]
table {%
1 0.0162053902943929
3 0.0395387490590413
5 0.0624022881189982
7 0.0829610347747803
9 0.106836024920146
11 0.129907854398092
13 0.151692040761312
};
\addplot [thin, color3, mark=*, mark size=1, mark options={solid}]
table {%
1 0.0224481821060181
3 0.0588473796844482
5 0.0956717014312744
7 0.131986101468404
9 0.1628178358078
11 0.201059206326803
13 0.235243622461955
};
\addplot [thin, color4, mark=*, mark size=1, mark options={solid}]
table {%
1 0.0287069082260132
3 0.0783005952835083
5 0.125459829966227
7 0.174980886777242
9 0.223523219426473
11 0.276958545049032
13 0.323186635971069
};
\addplot [thin, color5, dotted, mark=*, mark size=1, mark options={solid}]
table {%
1 0.0137713670730591
3 0.0146248022715251
5 0.0153758525848389
7 0.0162443161010742
9 0.0157034158706665
11 0.0159784475962321
13 0.0172077576319377
};
\end{axis}

\end{tikzpicture}%\hfill 
\hspace*{-1mm} 
% This file was created by matplotlib2tikz v0.6.7.
\begin{tikzpicture}[font=\tiny]

\definecolor{color1}{rgb}{1,0.498039215686275,0.0549019607843137}
\definecolor{color0}{rgb}{0.12156862745098,0.466666666666667,0.705882352941177}
\definecolor{color3}{rgb}{0.83921568627451,0.152941176470588,0.156862745098039}
\definecolor{color2}{rgb}{0.172549019607843,0.627450980392157,0.172549019607843}
\definecolor{color5}{rgb}{0.549019607843137,0.337254901960784,0.294117647058824}
\definecolor{color4}{rgb}{0.580392156862745,0.403921568627451,0.741176470588235}

\begin{axis}[
xlabel={\tiny Order $r$},
ylabel={\tiny Time Ratios },
xmin=0.4, xmax=13.6,
ymin=-2.3081905092183, ymax=24.9305032686226,
width=\fwidth,
height=\fheight,
tick align=outside,
%tick pos=left,
yticklabel pos=right,
x grid style={lightgray!92.026143790849673!black},
y grid style={lightgray!92.026143790849673!black},
xtick={1, 3, 5, 7, 9, 11, 13},
ytick={0, 3, 6, 9, 12, 15, 18, 21, 24},
%legend entries={{Matrix inversion},{No. Jacobi iter 1},{No. Jacobi iter 5},{No. Jacobi iter 9},{No. Jacobi iter 13},{ChebNet}},
%legend cell align={left},
%legend style={at={(0.03,0.97)}, anchor=north west, draw=white!80.0!black}
]
\path [draw=color0, thin] (axis cs:1,10.9328087368348)
--(axis cs:1,14.1528524238385);

\path [draw=color0, thin] (axis cs:3,10.4251526210235)
--(axis cs:3,13.2627444090152);

\path [draw=color0, thin] (axis cs:5,9.89633475553664)
--(axis cs:5,12.8074058180835);

\path [draw=color0, thin] (axis cs:7,9.30556937098507)
--(axis cs:7,12.2901239922503);

\path [draw=color0, thin] (axis cs:9,10.2646389205534)
--(axis cs:9,13.172316625319);

\path [draw=color0, thin] (axis cs:11,10.1671777774736)
--(axis cs:11,12.7988485018504);

\path [draw=color0, thin] (axis cs:13,9.12893725133587)
--(axis cs:13,12.4456608180672);

\path [draw=color1, thin] (axis cs:1,-0.474858873508567)
--(axis cs:1,2.38028690230986);

\path [draw=color1, thin] (axis cs:3,0.0422182762301846)
--(axis cs:3,2.48327044091651);

\path [draw=color1, thin] (axis cs:5,-0.737303212061811)
--(axis cs:5,4.51608474746985);

\path [draw=color1, thin] (axis cs:7,-0.475013927081148)
--(axis cs:7,5.08554187022031);

\path [draw=color1, thin] (axis cs:9,-0.82520490748907)
--(axis cs:9,7.19675738333989);

\path [draw=color1, thin] (axis cs:11,-1.06226806729995)
--(axis cs:11,8.12681346125605);

\path [draw=color1, thin] (axis cs:13,-1.07006806477099)
--(axis cs:13,8.9058911178594);

\path [draw=color2, thin] (axis cs:1,-0.0356704162679002)
--(axis cs:1,2.38916084440346);

\path [draw=color2, thin] (axis cs:3,1.37949117656937)
--(axis cs:3,4.0275903449394);

\path [draw=color2, thin] (axis cs:5,1.68892681689819)
--(axis cs:5,6.42799389038397);

\path [draw=color2, thin] (axis cs:7,1.77731658783744)
--(axis cs:7,8.43684500057121);

\path [draw=color2, thin] (axis cs:9,2.63457977059613)
--(axis cs:9,10.9721444988301);

\path [draw=color2, thin] (axis cs:11,3.69201379932437)
--(axis cs:11,12.5683711493017);

\path [draw=color2, thin] (axis cs:13,2.61162355473323)
--(axis cs:13,15.0190339668222);

\path [draw=color3, thin] (axis cs:1,0.880994282280626)
--(axis cs:1,2.37912971077355);

\path [draw=color3, thin] (axis cs:3,2.15753294848224)
--(axis cs:3,5.89008076852861);

\path [draw=color3, thin] (axis cs:5,3.3820653027157)
--(axis cs:5,9.06234399470564);

\path [draw=color3, thin] (axis cs:7,5.2816343324804)
--(axis cs:7,10.9684928685873);

\path [draw=color3, thin] (axis cs:9,6.21516486539356)
--(axis cs:9,14.5214490215174);

\path [draw=color3, thin] (axis cs:11,7.41272654945104)
--(axis cs:11,17.753573883166);

\path [draw=color3, thin] (axis cs:13,8.73343445807715)
--(axis cs:13,18.6081433922921);

\path [draw=color4, thin] (axis cs:1,0.673314613293587)
--(axis cs:1,3.49575706618731);

\path [draw=color4, thin] (axis cs:3,4.16050075399923)
--(axis cs:3,6.54741772992611);

\path [draw=color4, thin] (axis cs:5,6.36510728943431)
--(axis cs:5,9.95396565614466);

\path [draw=color4, thin] (axis cs:7,8.37183352132729)
--(axis cs:7,13.1718111219422);

\path [draw=color4, thin] (axis cs:9,10.5384844068731)
--(axis cs:9,17.9296172173102);

\path [draw=color4, thin] (axis cs:11,10.9741339772177)
--(axis cs:11,23.6923808241753);

\path [draw=color4, thin] (axis cs:13,14.2470625414027)
--(axis cs:13,23.3158370395669);

\path [draw=color5, thin] (axis cs:1,0.824772116146981)
--(axis cs:1,1.17522788385302);

\path [draw=color5, thin] (axis cs:3,0.830261581636613)
--(axis cs:3,1.16973841836339);

\path [draw=color5, thin] (axis cs:5,0.770168530726767)
--(axis cs:5,1.22983146927323);

\path [draw=color5, thin] (axis cs:7,0.591175506323512)
--(axis cs:7,1.40882449367649);

\path [draw=color5, thin] (axis cs:9,0.786692908570382)
--(axis cs:9,1.21330709142962);

\path [draw=color5, thin] (axis cs:11,0.757752489460665)
--(axis cs:11,1.24224751053934);

\path [draw=color5, thin] (axis cs:13,0.530231842943414)
--(axis cs:13,1.46976815705659);

\addplot [thin, color0, mark=-, mark size=3, mark options={solid}, only marks]
table {%
1 10.9328087368348
3 10.4251526210235
5 9.89633475553664
7 9.30556937098507
9 10.2646389205534
11 10.1671777774736
13 9.12893725133587
};
\addplot [thin, color0, mark=-, mark size=3, mark options={solid}, only marks]
table {%
1 14.1528524238385
3 13.2627444090152
5 12.8074058180835
7 12.2901239922503
9 13.172316625319
11 12.7988485018504
13 12.4456608180672
};
\addplot [thin, color1, mark=-, mark size=3, mark options={solid}, only marks]
table {%
1 -0.474858873508567
3 0.0422182762301846
5 -0.737303212061811
7 -0.475013927081148
9 -0.82520490748907
11 -1.06226806729995
13 -1.07006806477099
};
\addplot [thin, color1, mark=-, mark size=3, mark options={solid}, only marks]
table {%
1 2.38028690230986
3 2.48327044091651
5 4.51608474746985
7 5.08554187022031
9 7.19675738333989
11 8.12681346125605
13 8.9058911178594
};
\addplot [thin, color2, mark=-, mark size=3, mark options={solid}, only marks]
table {%
1 -0.0356704162679002
3 1.37949117656937
5 1.68892681689819
7 1.77731658783744
9 2.63457977059613
11 3.69201379932437
13 2.61162355473323
};
\addplot [thin, color2, mark=-, mark size=3, mark options={solid}, only marks]
table {%
1 2.38916084440346
3 4.0275903449394
5 6.42799389038397
7 8.43684500057121
9 10.9721444988301
11 12.5683711493017
13 15.0190339668222
};
\addplot [thin, color3, mark=-, mark size=3, mark options={solid}, only marks]
table {%
1 0.880994282280626
3 2.15753294848224
5 3.3820653027157
7 5.2816343324804
9 6.21516486539356
11 7.41272654945104
13 8.73343445807715
};
\addplot [thin, color3, mark=-, mark size=3, mark options={solid}, only marks]
table {%
1 2.37912971077355
3 5.89008076852861
5 9.06234399470564
7 10.9684928685873
9 14.5214490215174
11 17.753573883166
13 18.6081433922921
};
\addplot [thin, color4, mark=-, mark size=3, mark options={solid}, only marks]
table {%
1 0.673314613293587
3 4.16050075399923
5 6.36510728943431
7 8.37183352132729
9 10.5384844068731
11 10.9741339772177
13 14.2470625414027
};
\addplot [thin, color4, mark=-, mark size=3, mark options={solid}, only marks]
table {%
1 3.49575706618731
3 6.54741772992611
5 9.95396565614466
7 13.1718111219422
9 17.9296172173102
11 23.6923808241753
13 23.3158370395669
};
\addplot [thin, color5, mark=-, mark size=3, mark options={solid}, only marks]
table {%
1 0.824772116146981
3 0.830261581636613
5 0.770168530726767
7 0.591175506323512
9 0.786692908570382
11 0.757752489460665
13 0.530231842943414
};
\addplot [thin, color5, mark=-, mark size=3, mark options={solid}, only marks]
table {%
1 1.17522788385302
3 1.16973841836339
5 1.22983146927323
7 1.40882449367649
9 1.21330709142962
11 1.24224751053934
13 1.46976815705659
};
\addplot [thin, color0, dashed, mark=*, mark size=1, mark options={solid}]
table {%
1 12.5428305803367
3 11.8439485150194
5 11.35187028681
7 10.7978466816177
9 11.7184777729362
11 11.483013139662
13 10.7872990347015
};
\addplot [thin, color1, mark=*, mark size=1, mark options={solid}]
table {%
1 0.952714014400645
3 1.26274435857335
5 1.88939076770402
7 2.30526397156958
9 3.18577623792541
11 3.53227269697805
13 3.91791152654421
};
\addplot [thin, color2, mark=*, mark size=1, mark options={solid}]
table {%
1 1.17674521406778
3 2.70354076075439
5 4.05846035364108
7 5.10708079420433
9 6.80336213471313
11 8.13019247431305
13 8.81532876077769
};
\addplot [thin, color3, mark=*, mark size=1, mark options={solid}]
table {%
1 1.63006199652709
3 4.02380685850542
5 6.22220464871067
7 8.12506360053385
9 10.3683069434555
11 12.5831502163085
13 13.6707889251846
};
\addplot [thin, color4, mark=*, mark size=1, mark options={solid}]
table {%
1 2.08453583974045
3 5.35395924196267
5 8.15953647278949
7 10.7718223216347
9 14.2340508120916
11 17.3332574006965
13 18.7814497904848
};
\addplot [thin, color5, dotted, mark=*, mark size=1, mark options={solid}]
table {%
1 1
3 1
5 1
7 1
9 1
11 1
13 1
};
\end{axis}

\end{tikzpicture}%\hfill
\end{minipage}
\begin{minipage}{1.0\linewidth}
\input{figures/community_detection/new_times/CayleyNet_train_time_inv_jacobi_order_batch_analysis_multiple_size_max_1000vert_sparse_with_std_neuro.tex}%\hfill 
\hspace*{-2.6mm} 
\input{figures/community_detection/new_times/CayleyNet_train_time_inv_jacobi_order_batch_analysis_multiple_size_max_1000vert_sparse_ratio_with_std_neuro.tex}%\hfill
\end{minipage}
\vspace{-2mm}
\end{minipage}
%}
%\input{../../figures/community_detection/CayleyNet_test_time_inv_jacobi_order_batch_analysis_multiple_size_max_1000vert_sparse_iclr.tex}\hfill
%\vspace{-2mm}
\vspace{2mm}
\caption{Test (above) and training (below) times with corresponding ratios as function of filter order $r$ and graph size $n$ on our community detection dataset.}
\label{fig:test-training-times-community-dataset}
\end{figure}

\section{{Back propagation}}

In this section we show how to differentiate Cayley filters $\mathbf{G}_{\mathbf{c},h}=g_{c,h}(\boldsymbol{\Delta})$ with respect to the complex coefficient vector $\mathbf{c}=(c_0,\ldots,c_r)$ and $h$. Since working with complex parameters is not standard, we explain in detail how this is done. One approach is to simply treat each complex parameter $c_j=c_j^R+ic_j^I$ as the pair of real parameters $(c_j^R,c_j^I)$, and to explicitly formulate Cayley polynomial with real numbers. This brute force formulation is suitable for automatic back propagation in software like TensorFlow. To justify the calculation from a theoretical standpoint, it is more convenient to consider a general calculus of variation approach to gradient descent. 

Our goal is to minimize the loss function with respect to all of the coefficients of all filters in the network. 
Note that minimization is a set operation in its nature. Namely, a minimal value of a set doesn't depend on additional structures endowed on the set, such as inner product, topology, Riemannian structure, and so on. This means that we are free to use the vector space structure and the inner product of our choice to define the gradient. 

Consider a generic Cayley filter, applied on the real valued signal $\mathbf{f}$
\[\mathbf{G}_{\mathbf{c},h}\mathbf{f}= c_0\mathbf{f} + \sum_{j=1}^{r} 2\Re \{c_j\cC^{j}(h \boldsymbol{\Delta})\mathbf{f}\}.\]
Let 
\begin{equation}
S(\mathbf{c},h)=F(\mathbf{G}_{\mathbf{c},h}\mathbf{f})
\label{eq:prop1}
\end{equation} denote the dependency of the loss function on $(\mathbf{c},h)$. %, where $\mathbf{f}$ is one input data at the depth corresponding to the filter $p(c;L)$.
 Our goal is to define an inner product, and calculate a gradient of $S(\mathbf{c},h)$ with respect to the complex coefficient vector $\mathbf{a}=(\mathbf{c},h)$.
Given an inner product structure on the space of coefficients, a (variational) gradient at the point $\mathbf{a}$ of a scalar valued function $S$ is a vector $D[S;\mathbf{a}]$ such that for any other coefficient vector $\boldsymbol{\eta}$
\[S(\mathbf{a}+\epsilon\boldsymbol{\eta}) - S(\mathbf{a}) = \epsilon\left\langle{D[S;\mathbf{a}]},{\boldsymbol{\eta}}\right\rangle + o(\epsilon)\]
where $\epsilon$ denotes a scalar. If there is no such vector $D[S;\mathbf{a}]$, $S$ is not differentiable at $\mathbf{a}$ by definition.
This definition can be extended to vector valued functions.

It can be shown that under the standard complex dot product of the coefficient space, $S$ is not differentiable in general.
However, by defining a new inner product in the coefficient space, there is a way to make $S$ differentiable. 
For intuition, consider the vector space $\CC$ with the classical inner product $\left\langle z,w\right\rangle=z\overline{w}$.  Here, the space of differentiable functions $S:\CC\rightarrow\CC$ are the analytic functions. However, if we define the new inner product $\Re\left\langle z,w\right\rangle= z_Rw_R + z_Iw_I$, $\CC$ is isometrically isomorphic to $\RR^2$ with the standard real inner product. Moreover, if we treat the image space $\CC$ of $S$ as $\RR^2$, the space of differentiable functions $S:\RR^2\rightarrow\RR^2$ is the richer space of classical real differentiable functions. This procedure of defining a real vector space from a complex one is called realification (\cite{constantin_2016} page 117). 

Given a general complex Hilbert space $\mathcal{H}$, it's \textbf{realification} is the real Hilbert space $\mathcal{H}_{\RR}$ defined to be $\mathcal{H}$ restricted to multiplication by real scalars. The inner product of two vectors $\mathbf{f},\mathbf{g}$ in $\mathcal{H}_{\RR}$ is defined to be the real part of the inner product in $\mathcal{H}$, $\Re\left\langle \mathbf{f},\mathbf{g}\right\rangle$. 

In our case, we treat the coefficient space as $\RR^{2r}$. Indeed, $h$ and $c_0$ are real, and the rest of the coefficients are complex. 
Let us differentiate $\mathbf{G}_{\mathbf{c},h}\mathbf{f}$ with respect to $\mathbf{c}$ by definition.
\[\frac{\mathbf{G}_{\mathbf{c}+\epsilon\boldsymbol{\eta},h}\mathbf{f} - \mathbf{G}_{\mathbf{c},h}\mathbf{f}}{\epsilon} =  \mathbf{G}_{\boldsymbol{\eta},h}\mathbf{f}\]
\[=\Re\big\{\eta_0 \mathbf{f} + \sum_{j=1}^{r} 2\eta_j\cC^{j}(h \boldsymbol{\Delta})\mathbf{f}\big\}.\]
Thus, by the definition of the inner product in the (realificated) coefficient space, the differential of $\mathbf{G}_{\mathbf{c},h}\mathbf{f}$ with respect to $\mathbf{c}$ is given by the vector $D_{\mathbf{c}}[\mathbf{G}_{\mathbf{c},h}\mathbf{f};(h,\mathbf{c})]$ with vector valued entries
\[D_{\mathbf{c}}[\mathbf{G}_{\mathbf{c},h}\mathbf{f};(h,\mathbf{c})]_j = 
\left\{
\begin{array}{ccc}
	\mathbf{f} & , & j=0 \\
	2\cC^{j}(h \boldsymbol{\Delta})\mathbf{f} & , & {\rm else}
\end{array}  
\right.\]

Next we show that back propagation works in the realificated space $\RR^{2r}$ in the usual way.
By (\ref{eq:prop1}) and by the chain rule, we can write up to $o(\epsilon)$
\[\frac{S(\mathbf{c}+\epsilon\boldsymbol{\eta}) - S(\mathbf{c})}{\epsilon} \approx \nabla F(\mathbf{G}_{\mathbf{c},h}\mathbf{f}) \cdot \Re\left\{D_{\mathbf{c}}[\mathbf{G}_{\mathbf{c},h}\mathbf{f}]\boldsymbol{\eta}\right\}\]
where $\nabla$ denotes the gradient with respect to the signal, and $\cdot$ is the usual real dot product between real valued signals. Let us write in short $\nabla F=\nabla F(\mathbf{G}_{\mathbf{c},h}\mathbf{f})$, and obtain
\begin{multline}
\frac{S(\mathbf{c}+\epsilon\mathbf{\eta}) - S(\mathbf{c})}{\epsilon} \\
\approx  \Re\left\{  \nabla F \cdot    \left(\eta_0 \mathbf{f} + \sum_{j=1}^{r} 2\eta_j\cC^{j}(h \boldsymbol{\Delta})\mathbf{f}\right)\right\} \\
       = \Re\left\{  [\nabla F \cdot \mathbf{f}] \eta_0 + \sum_{j=1}^{r} 2 [\nabla F \cdot \cC^{j}(h \boldsymbol{\Delta})\mathbf{f}] \eta_j \right\} \\
			= \Re\left\langle {\nabla F \cdot D_{\mathbf{c}}[\mathbf{G}_{\mathbf{c},h}\mathbf{f};(h,\mathbf{c})]}\ ,\ {\mathbf{\eta}}\right\rangle .
\end{multline}
This shows, by the definition of the inner product in the (realificated) coefficient space, that
\[\nabla S = \nabla F \cdot D[p;(\mathbf{c},f)].\]
Namely, the chain rule works in the usual way in the realificated space $\RR^{2r}$.

Next we calculate the partial derivative with respect to the spectral zoom $h$. We start by standard Calculus on the function $g_{\mathbf{c},h}(\lambda)$, $\lambda\in\RR$, and get 
\[\frac{\partial}{\partial h}g_{\mathbf{c},h}(\lambda) = \sum_{j=1}^r c_j j \cC^{j-1}(h\lambda)\cC'(h\lambda)\lambda - \overline{c_j} j \cC^{-j-1}(h\lambda)\cC'(h\lambda)\lambda \]
where $\cC'(x) = (x+i)^{-1}(1-\cC(x))$.
We can interpret this as a calculation on the spectrum of $\mathbf{G}_{\mathbf{c},h}=g_{\mathbf{c},h}(\boldsymbol{\Delta})$. Then, by the fact that $\boldsymbol{\Delta}$ is a bounded normal operator, we can carry the calculation to $\mathbf{G}_{\mathbf{c},h} = g_{\mathbf{c},h}(\boldsymbol{\Delta})$ via functional calculus. We thus obtain
\[
D_{h}[\mathbf{G}_{\mathbf{c},h}\mathbf{f};(h,\mathbf{c})]
= \cC'(h\boldsymbol{\Delta})\boldsymbol{\Delta}\sum_{j=1}^r jc_j  \Big(\cC^{j-1}(h\boldsymbol{\Delta}) - j\overline{c_j}  \cC^{-j-1}(h\boldsymbol{\Delta})\Big)\mathbf{f} , 
\]
where $\cC'(h\boldsymbol{\Delta}) = (h\boldsymbol{\Delta}+i\mathbf{I})^{-1}(\mathbf{I}-\cC(h\boldsymbol{\Delta}))$.

\section*{Acknowledgment}

MB and FM are partially supported by ERC Consolidator Grant No. 724228 (LEMAN), Google Faculty Research Awards, Amazon AWS ML Research Award, Royal Society Wolfson Research Merit Award, and Rudolf Diesel fellowship at the Institute for Advanced Studies, TU Munich.

\bibliographystyle{abbrv}
\bibliography{IEEEabrv,IEEE_Cayley}

\end{document}